\documentclass[]{article}
\pdfoutput=1

\usepackage[preprint,nonatbib]{neurips_2024}
\usepackage[utf8]{inputenc} % allow utf-8 input
\usepackage[T1]{fontenc}    % use 8-bit T1 fonts
\usepackage{color, colortbl}
\usepackage[dvipsnames]{xcolor}
\definecolor{citecolor}{HTML}{2980b9}
\definecolor{linkcolor}{HTML}{2980b9}
\definecolor{urlcolor}{HTML}{483D8B}
\definecolor{notecolor}{HTML}{6495ED}
\definecolor{notecolor2}{HTML}{4682B4}
\usepackage{hyperref}

\usepackage{url}            % simple URL typesetting
\usepackage{booktabs}       % professional-quality tables
\usepackage{amsfonts}       % blackboard math symbols
\usepackage{nicefrac}       % compact symbols for 1/2, etc.
\usepackage{microtype}      % microtypography
\usepackage{times}
\usepackage{epsfig}
\usepackage{graphicx}
\usepackage{grffile}
\usepackage{amssymb}
\usepackage{amsmath}
\usepackage{cleveref}
\usepackage{romannum}
\usepackage{threeparttable}

\DeclareMathOperator*{\argmax}{arg\,max}

\usepackage{multirow} % used by tables
\usepackage{makecell} % used by tables
\usepackage{diagbox} % used in the table with the backslashbox
\usepackage{siunitx}

\usepackage{enumitem} % used for itemize margins
\usepackage{arydshln} % dashed lines in tables
\usepackage{tabularx}
\usepackage{xspace} % deal with spacing after newcommand
\usepackage[subrefformat=parens,labelformat=parens]{subcaption} % provides subfigure
\usepackage[Export]{adjustbox} % provides valign of images and align
\usepackage{pifont} % for red cross
\usepackage{balance} % balance bibliography
\usepackage{wrapfig}
\usepackage[linesnumbered,ruled]{algorithm2e}

\crefname{table}{Table.}{Tables.}
\crefname{figure}{Figure.}{Figures.}
\crefname{section}{Section.}{Sections.}

\newcommand{\ours}{LiveScene\xspace}
\newcommand{\simdata}{OmniSim\xspace}
\newcommand{\realdata}{InterReal\xspace}

\newcounter{boldpara}
\newcommand{\boldparagraph}[1]{
    \refstepcounter{boldpara}
    \vspace{0.1em}\noindent{\bf #1}
}

\newcommand{\blackcheck}{{\color{black}\checkmark}}
\definecolor{mblue}{HTML}{367dbd}
\newcommand{\bluecheck}{{\textcolor{mblue}{\checkmark}}}

\newcommand{\blackx}{{\color{black}\ding{55}}}

\colorlet{colorFst}{Gray!45}       % first
\colorlet{colorSnd}{Gray!25} % second

\colorlet{colorTrd}{yellow!30}      % third
\colorlet{colorLow}{darkgray!30}    % low-light color
\definecolor{R1}{HTML}{E97451}
\definecolor{R2}{HTML}{008080}
\definecolor{R3}{HTML}{0047AB}
\colorlet{cmt}{darkgray!80}    % low-light color
\colorlet{supp}{darkgray!50}    % low-light color

\newcommand{\fs}{\cellcolor{colorFst}\bf}   % first
\newcommand{\nd}{\cellcolor{colorSnd}}      % second
\let\titleold\title
\renewcommand{\title}[1]{\titleold{#1}\newcommand{\thetitle}{#1}}
\def\maketitlesupplementary
{
    \newpage
    \begin{center}
        \Large
        \textbf{\thetitle}\\
        \vspace{0.5em}Supplementary Material \\
        \vspace{1.0em}
    \end{center}
}
\title{LiveScene: Language Embedding Interactive Radiance Fields for Physical Scene Rendering and Control}

\author{
        Delin Qu$^{1,2}$\thanks{Authors contributed equally: \href{mailto:dlqu22@m.fudan.edu.cn}{dlqu22@m.fudan.edu.cn}.}  \hspace{3.5em} 
        Qizhi Chen$^{3,2}$\footnotemark[1] \hspace{1em} 
        Pingrui Zhang$^{1,2}$ \hspace{1em}
        Xianqiang Gao$^{2}$ \hspace{1em} 
        Junzhe Li$^{4}$ \\
        \textbf{Bin Zhao}$^{2}$ \hspace{1em}
        \textbf{Zhigang Wang}$^{2}$ \hspace{1em}
        \textbf{Dong Wang}$^{2}$\thanks{Corresponding author: \href{mailto:dongwang.dw93@gmail.com}{dongwang.dw93@gmail.com}.} \hspace{1em}
        \textbf{Xuelong Li}$^{2}$ \\
        $^{1}$Fudan University \hspace{0em}
        $^{2}$Shanghai AI Laboratory \hspace{0em}
        $^{3}$Zhejiang University \hspace{0em}
        $^{4}$Peking University \hspace{0em}
}
\begin{document}
\pagenumbering{arabic}

\maketitle
\begin{figure}[htbp]
    \vspace{-3ex}
    \begin{center}
        \includegraphics[width=1.0\linewidth]{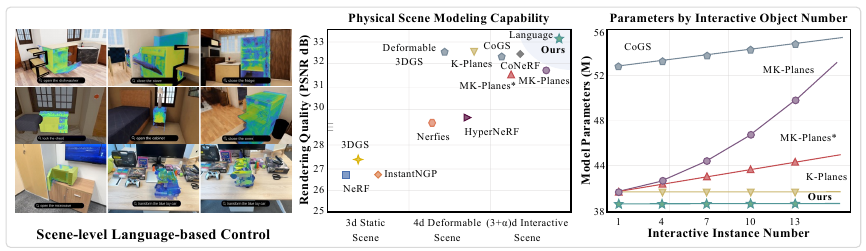}
    \end{center}
    \vspace{-2ex}
    \caption{\ours enables scene-level reconstruction and control with language grounding. Left: Language-interactive articulated object control in Nerfstudio. Right: \ours achieves SOTA rendering quality on \simdata dataset and exhibits a significant advantage in parameter efficiency.}
    \label{fig:head}
    % \vspace{-5ex}
\end{figure}
\begin{abstract}
    % This paper aims to advance the progress of interactive scene reconstruction by scaling up object-level reconstruction to complex scenes. Specifically, we introduce \simdata and \realdata datasets comprising 28 scenes with multiple interactive objects. To overcome the challenge of inaccuracy of interactive motion recovery in complex scenes, we propose \textbf{\ours}, the first scene-level language-embedded interactive radiance field that efficiently reconstructs and controls multiple objects in complex scenes. \ours leverages an efficient factorization that decomposes the interactive scene into local deformable fields, enabling separate reconstruction of individual object motions and significantly reducing memory consumption. Moreover, we propose an interaction-aware language embedding for individual interactive object localization, enabling arbitrary control of interactive objects using natural language. Extensive experiments demonstrate our method’s effectiveness and efficiency with significant superiority in novel view synthesis and language grounding performance. Project page: \href{https://livescenes.github.io}{https://livescenes.github.io}.
    This paper scales object-level reconstruction to complex scenes, advancing interactive scene reconstruction. We introduce two datasets, \simdata and \realdata, featuring 28 scenes with multiple interactive objects. To tackle the challenge of inaccurate interactive motion recovery in complex scenes, we propose \textbf{\ours}, a scene-level language-embedded interactive radiance field that efficiently reconstructs and controls multiple objects. By decomposing the interactive scene into local deformable fields, \ours enables separate reconstruction of individual object motions, reducing memory consumption. Additionally, our interaction-aware language embedding localizes individual interactive objects, allowing for arbitrary control using natural language. Our approach demonstrates significant superiority in novel view synthesis, interactive scene control, and language grounding performance through extensive experiments. Project page: \href{https://livescenes.github.io}{https://livescenes.github.io}.
\end{abstract}
\section{Introduction}
\label{sec:intro}
Interactive objects are prevalent in our daily lives, and modeling interactable scenes from the real physical world plays an essential role in various research fields, including content generation~\cite{liao2024advances,liu2023zero1to3,tang2023dreamgaussian}, animation~\cite{Wang2023AdaptiveSF,luo2024gaussianhair,li2022tava}, virtual reality~\cite{Steuer1992DefiningVR,Flavin2019TheIO,Liao_2023_CVPR}, robotics~\cite{Qu_2024_CVPR,zhou2024smartrefine,zhou2025smartpretrain,liu2024coherentcollaborationheterogeneousmultirobot,qu2025spatialvlaexploringspatialrepresentations}, and world understanding~\cite{jing2024x4d,guo2023viewrefer,tang2024any2point,jing2022towards,gao2024aerial}. This paper tackles the challenging and rarely explored task of reconstructing and controlling multiple interactive objects in complex scenes from a single, casually captured monocular video without previous independent modeling of geometry and kinematics.
% \begin{figure}[t]
%     \vspace{-5ex}
%     \begin{center}
%         \includegraphics[width=0.98\linewidth]{./figures/head.pdf}
%     \end{center}
%     \vspace{-2ex}
%     \caption{Scene modeling capability, render performance, and efficiency comparisons among various methods. In contrast to static or deformable scenes, \ours is capable of modeling $(3+\alpha)$d interactive scenes with language grounding and achieving SOTA rendering quality on \simdata dataset. Notably, \ours exhibits a significant advantage in parameter efficiency, maintaining a constant complexity regardless of interactive instance number and satisfying rendering time at $1200 \times 1200$ resolution. In contrast, other methods, such as CoGS~\cite{yu2023cogs} and MK-Planes~\cite{kplanes_2023}, exhibit a linear or quadratic increase in parameter quantity.}
%     \label{fig:head}
%     \vspace{-5ex}
% \end{figure}
Prior research on interactable scene modeling, such as CoNeRF~\cite{conerf_kania_2022} and K-Planes~\cite{kplanes_2023}, typically adopts a joint modeling approach, combining spatial coordinates and all interaction variables as input and representing interactive scene by either implicit MLPs or feature planes. Meanwhile, CoGS~\cite{yu2023cogs} learns parameter offsets for different scene parts using multiple independent MLPs after establishing a 3D deformable Gaussian scene. However, these methods primarily focus on capturing interactions for a single object within a clear background, such as a single drawer, toy car, or face~\cite{yunus2024recent,conerf_kania_2022,yu2023cogs,zheng2023editablenerf}. As modeling extends from single objects to multiple objects in complex scenes, as shown in~\cref{fig:head}, the interaction spaces become increasingly high-dimensional, complicating these methods for accurate modeling and significantly increasing computational time and memory cost, e.g., 4$\times$ A100 GPU for 2 weeks to converge training in CoNeRF~\cite{conerf_kania_2022} and 500M Gaussian storage for a regular indoor living room in CoGS~\cite{yu2023cogs}. Moreover, natural language is an intuitive and necessary interface for interacting with 3D scenes, but language embedding of interactive scenes faces an even more daunting challenge: interaction variation inconsistency. For instance, methods like LERF~\cite{lerf_kerr_2023}, and OpenNeRF~\cite{zhang2024open}, which distill CLIP features into static 3D fields, suffer from significant failures when confronted with scene topology structure changes induced by interactions, such as the distinct structures variation of a cabinet before and after opening.

To address these challenges, we propose \ours, the first scene-level language-embedded radiance fields, which compresses high-dimensional interaction spaces into compact 4D feature planes, reducing model parameters while improving optimization effectiveness. \ours models multiple object interactions via novel high-dimensional factorization, decomposing the scene into local deformable fields that model individual objects with multi-scale 4D deformable feature planes. To achieve independent control, we introduce a multi-scale interaction probability sampling strategy for factorized local deformable fields. Our interaction-aware language embedding method generates varying language embeddings to localize and control objects under arbitrary states, enabling natural language control. Finally, we construct the first scene-level physical interaction datasets, \simdata and \realdata, featuring 28 scenes with 70 interactive objects for evaluation. 

Experiment results show that our approach achieves SOTA novel view synthesis quality, outperforming existing best methods by +9.89, +1.30, and +1.99 in PSNR on the CoNeRF Synthetic, \simdata \#chanllenging, and \realdata \#chanllenging subsets, respectively. Surpassing LeRF~\cite{lerf_kerr_2023}, \ours significantly improves language grounding accuracy by +65.12 of mIOU on the \simdata dataset. Notably, our method maintains a lightweight, constant model parameter of 39M, scaling well with increasing scene complexity, as shown in~\cref{fig:head}. Contributions can be summarized as:
\begin{itemize}[noitemsep, topsep=10pt, leftmargin=10pt]
    \item We propose \ours, the first scene-level language-embedded interactive radiance field, which efficiently reconstructs and controls complex physical scenes, enabling manipulation of multiple articulated objects and language-based interaction.
    \item We propose a factorization technique that decomposes interactive scenes into local deformable fields and samples relevant 3D points, enabling control of individual objects. Additionally, we introduce an interaction-aware language embedding method that generates varying embeddings, allowing for language-based control and localization.
    \item We construct the first scene-level physical interaction dataset \textbf{\simdata} and \textbf{\realdata}, containing 28 subsets and 70 interactive objects for evaluation. Extensive experiments demonstrate our SOTA performance and robust interaction capabilities.
\end{itemize}
\section{Related Work}
\label{sec:relate}
\boldparagraph{Dynamic Scene Representation.} 
Extending NeRF~\cite{mildenhall2020nerf} to dynamic scene reconstruction has made significant progress. Related methods can generally be categorized into \textit{time-varying methods}, \textit{deformable-canonical methods}, and \textit{hybrid representation methods}. The time-varying methods~\cite{fang2022fast,park2021nerfies,tretschk2021non,yuan2021star} typically model the radiance field directly over time, but struggle to separate dynamic and static objects. Deformable-canonical methods~\cite{gao2021dynamic,li2022neural,park2021hypernerf,xian2021space} decouple dynamic deformable field and static canonical space, modeling 4D by warping points with deformable field to query the canonical features. However, these methods face challenges in scene topology changes~\cite{kplanes_2023}. Hybrid representation methods, on the other hand, have achieved high-quality reconstruction and fast rendering by utilizing time-space feature planes~\cite{shao2023tensor4d,kplanes_2023,Cao2023HexPlane}, 4D hash encoding~\cite{wang2023masked}, dynamic voxels~\cite{wang2023mixed}, or triple fields~\cite{song2023nerfplayer}. Recently, several works~\cite{luiten2023dynamic,yang2023deformable3dgs,wu20234dgaussians,kratimenos2024dynmf} have introduced 3D gaussians~\cite{kerbl3Dgaussians} into dynamic scene reconstruction, achieving high-quality real-time rendering speeds. However, these methods are limited to reconstructing dynamic scenes and lack the ability to control and understand interactive scenes.

\boldparagraph{3D Vision-language Fields.}
Vision-language foundational models~\cite{clip,dinov2_oquab_2023,kirillov2023segany} with strong generalizability and adaptability inspires numerous language embedded scene representation for 3D scene understanding~\cite{bommasani2021opportunities,comprehensive_zhou_2023}, such as open-vocabulary segmentation~\cite{3d_liu_2023,interactive_goel_2022,neural_tschernezki_2022,garfield_kim_2024}, 3D visual question answering~\cite{hong2023threedclr,chen2023ll3da,jia2024sceneverse,conceptfusion_jatavallabhula_2023}, and 3D language grounding~\cite{visual_huang_2023,clipfields_shafiullah_2022,openvocabulary_chen_2023,conceptfusion_jatavallabhula_2023}. LeRF~\cite{lerf_kerr_2023} is the first to achieve open-vocabulary 3D queries by combining CLIP~\cite{clip} and DINO~\cite{emerging_caron_2021} with NeRF through feature distillation. Open-NeRF~\cite{zhang2024open} introduces an integrate-and-distill paradigm and leverages hierarchical embeddings to address 2D knowledge-distilling issues from SAM. LEGaussians~\cite{shi2023language} and LangSplat~\cite{qin2023langsplat} integrate semantic features into 3D gaussians~\cite{kerbl3Dgaussians} and achieve precision language query and efficient rendering. However, these methods are limited to static scene understanding and fail to generalize when the interactive scene topology changes.

\boldparagraph{Controllable Scene Representation.} Manipulating reconstructed assets or neural fields is of significant importance for avatar and robotic tasks~\cite{firoozi2023foundation,voltemorph_garbin_2022,conerf_kania_2022,controlnerf_lazova_2023}. CoNeRF~\cite{conerf_kania_2022} pioneered this effort by extending HyperNeRF~\cite{park2021hypernerf} and introduce a ﬁne-grained controlable neural field with 2D attribute mask and value annotations. CoNFies~\cite{confies_yu_2023} proposes an automatic controllable avatar system and accelerates rendering by distilling. More recently, CoGS~\cite{yu2023cogs} leveraged 3D Gaussians~\cite{kerbl3Dgaussians} to achieve real-time control of dynamic scenes without requiring explicit control signals. However, these methods typically lack natural language interaction capabilities, relying solely on manual control. Furthermore, most works focus on single or few object interactions, disregarding the interaction between different scene parts, limiting their real-world applications.
\section{Methodology}
\label{sec:method}
\begin{figure}[t]
    \vspace{-2ex}
    \begin{center}
        \includegraphics[width=1\linewidth]{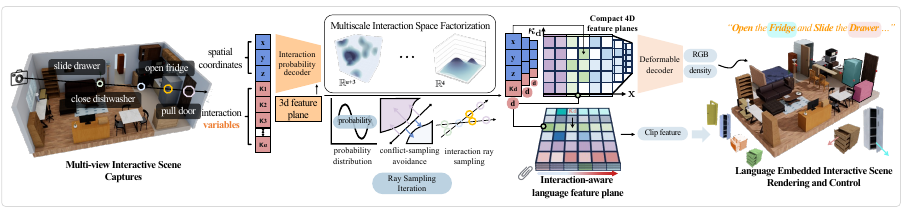}
    \end{center}
    \vspace{-1ex}
    \caption{The overview of \ours. Given a camera view and control variable $\boldsymbol{\kappa}$ of one specific interactive object, a series of 3D points are sampled in a local deformable field that models the interactive motions of this specific interactive object, and then the interactive object with novel interactive motion state is generated via volume-rendering. Moreover, an interaction-aware language embedding is utilized to localize and control individual interactive objects using natural language.}
    \label{fig:pipeline}
    \vspace{-2ex}
\end{figure}

We aim to establish a representation that models $\alpha$ interactive articulated objects in a complex scene from a monocular video via a rendering-based self-supervised manner. Control variables $\boldsymbol{\kappa} = \left [ \kappa_1, \kappa_2, ... , \kappa_\alpha \right ]$ indicating object motion states and camera poses of each video frame are given. The overview of \ours is shown in~\cref{fig:pipeline}. \cref{sec:interactive_space_modeling} introduces the high-dimensional interactive space modeling and challenges. \cref{sec:space_factor} presents a multi-scale interactive space factorization and sampling strategy to compress the high-dimensional interactive space into local 4D deformable fields, and model complicated interactive motions of individual objects. \cref{sec:language} introduces an interaction-aware language embedding method to localize and control interactive objects using natural language.

\subsection{Interactive Space}
\label{sec:interactive_space_modeling}
Assuming a non-rigidly interactive scene with $\alpha$ control variables $\boldsymbol{\kappa} = \left [ \kappa_1, \kappa_2, ... , \kappa_\alpha \right ]$ corresponding to $\alpha$ objects, we delineate its representation by a high-dimensional function:
\begin{equation}
    \mathbf{y}=\rho(\mathbf{x}, \boldsymbol{\kappa}, \mathcal{H} ; \boldsymbol{\theta}),
    \label{eq:model}
\end{equation}
where $\rho$ is the model of the representation, $\mathbf{x} \in \mathbb{R}^3$ are spatial coordinates, $\boldsymbol{\kappa} \in \mathbb{R}^\alpha$ are control variables, $\mathcal{H}$ is a set of optional additional parameters (e.g., the view direction), and $\boldsymbol{\theta}$ stores the scene information. The function outputs scene properties $\mathbf{y}$ for the given position $\mathbf{x}$ and $\boldsymbol{\kappa}$ sampling from a ray $\mathbf{r}$, where $\mathbf{y}$ can be represented with color, occupancy, signed distance, density, and BRDF parameters. This paper focuses on color, probability, and language embedding.  Distinguishing from 3d static scene or 4d dynamic scene modeling, the sampling point $\mathbf{p} = \left [ \mathbf{x} | \boldsymbol{\kappa} \right ] \in \mathbb{R}^{3+\alpha}$ in the interactive scene is high-dimensional and variable in topological structure, complicating the scene feature storage and the optimization of representation model, leading to significant time-consuming or memory-intensive training in~\cite{conerf_kania_2022,yu2023cogs}.

\subsection{Multi-scale Interaction Space Factorization}
\label{sec:space_factor}
\begin{wrapfigure}{r}{0.55\textwidth}
    \vspace{-5mm}
    \centering
    \includegraphics[width=0.55\textwidth]{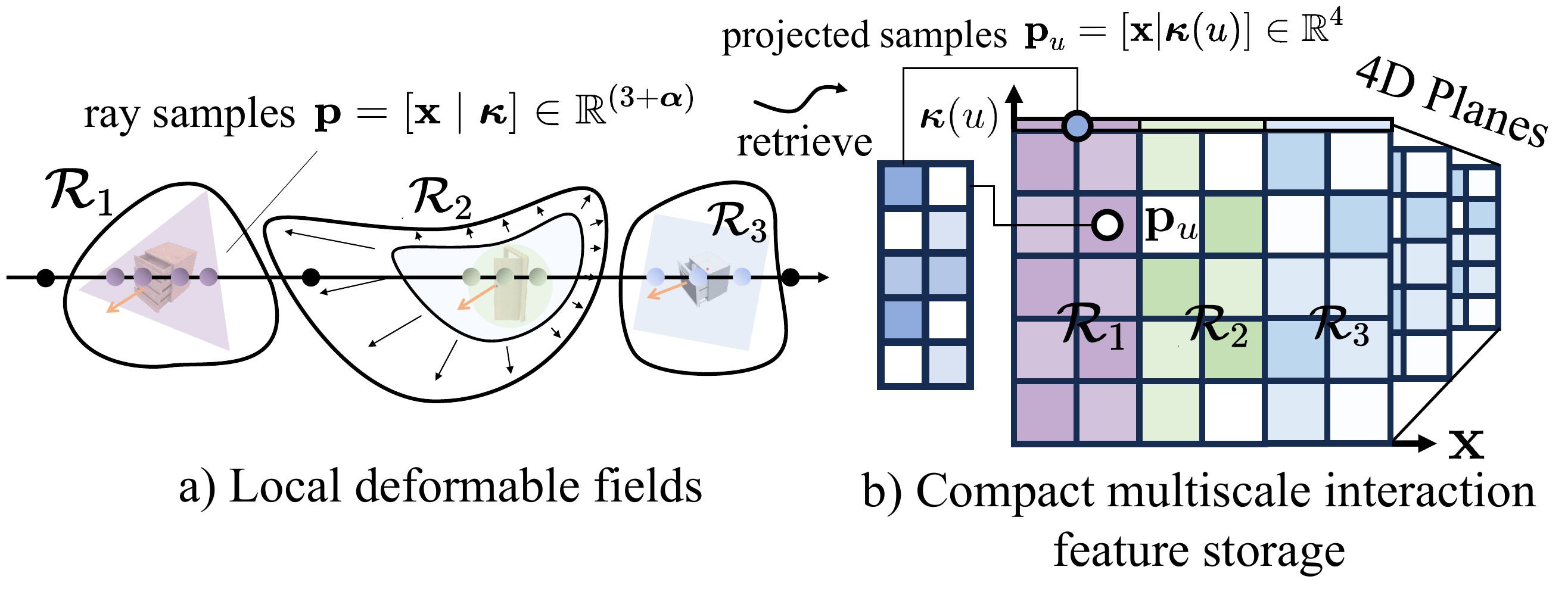}
    \vspace{-5mm}
    \caption{Illustration of hyperplanar factorization for compact storage. We maintain multiple local deformable fields for each interactive object region $\mathcal{R}_i$, and project high-dimensional interaction features into a compact 4D space, which can be further compressed into multiscale feature planes.}
    \label{fig:ray_sampling}
\end{wrapfigure}

For the $(3+\alpha)$-dimensional interactive space containing $\alpha$ control variables, we aim to explicitly represent the high-dimensional space in a concise and compact storage, thereby reducing memory usage and improving optimization. As illustrated in~\cref{fig:pipeline} and~\cref{fig:training}, objects exhibit mutual independence, and interaction features are distributed in the $(3+\alpha)$-dimensional interactive space and aggregate into cluster centers. Thus, there exists a set of hyperplanes that partition the space into disjoint regions, with each region containing a local 4D deformable field, as shown in~\cref{fig:ray_sampling}. Hence, the interaction features at sampling point $\mathbf{p} \in \mathbb{R}^{3+\alpha}$ can be projected into a compact 4-dimensional space $\mathbb{R}^{4}$ in~\cref{fig:ray_sampling} by a transformation.

\textbf{Multi-scale Interactive Ray Sampling.} We consider using ray sampling to perform the projection transformation. As shown in~\cref{fig:ray_sampling}, 
assuming a ray $\mathbf{r}(t) = \left [ \mathbf{o_x} | \mathbf{o}_{\boldsymbol{\kappa}} \right ] + t \left [ \mathbf{d_x} | \mathbf{d}_{\boldsymbol{\kappa}}  \right ]$ with origin $\mathbf{o_x} \in \mathbb{R}^{3}, \mathbf{o}_{\boldsymbol{\kappa}} \in \mathbb{R}^{\alpha}$, and direction $\mathbf{d_x} \in \mathbb{R}^{3}, \mathbf{d}_{\boldsymbol{\kappa}} \in \mathbb{R}^{\alpha}$, the ray intersects with the interaction region at a point $\mathbf{p} = \left [ \mathbf{x} | \boldsymbol{\kappa} \right ] \in \mathbb{R}^{3+\alpha}$, where $\mathbf{x} \in \mathbb{R}^{3}$ is the 3D position and $\boldsymbol{\kappa} \in \mathbb{R}^{\alpha}$ is the interaction variables. For a given intersection point $\mathbf{p}$, the deformable features can be retrieved from the corresponding local 4D deformable field by maximizing sampling probability $\mathbf{P}$:
\begin{equation}
    \mathbf{p}_u = \left [\mathbf{x} | \boldsymbol{\kappa}(u) \right ],\quad u = \argmax_{i} \{\mathbf{P}_i\}, \quad \mathbf{P} = \Theta(\boldsymbol{\kappa}, \boldsymbol{\theta}),
\end{equation}
where $\mathbf{p}_u$ and $\boldsymbol{\theta}$ are the 4D sampling point and probability features at position $\mathbf{x}$ from 3D feature planes. The maximum argument operation of probability decoder $\Theta(\boldsymbol{\kappa}, \boldsymbol{\theta})$ maps the interaction variables $\boldsymbol{\kappa}$ to the most probable cluster region in the 4D space, and can be optimized by minimizing the focal loss $\mathcal{L}_\text{focus}$ of mask across all the training camera views:
\begin{equation}
    \begin{aligned}
        \mathcal{L}_\text{focus} = \beta \cdot \left(1 - e^{\sum_{i=1}^{\alpha} \mathbf{M}_i \log(\hat{\mathbf{P}}_i)}\right)^{\gamma} \cdot \left(- \sum_{i=1}^{\alpha} \mathbf{M}_i \log(\hat{\mathbf{P}}_i)\right),
    \end{aligned}
\end{equation}
where $\mathbf{M}$ is the ground truth mask label, $\hat{\mathbf{P}}$ is the probability map rendering from the interactive probability field, $\beta$ is the balancing factor, and $\gamma$ is the focusing parameter. 

Next, the deformable features are used to render local deformable scene color at the sampling point $\mathbf{p}$. In this way, the high-dimensional interaction features are factorized into a 4D space by a lightweight transformation modeling with interaction probability decoder and 3D feature planes in~\cref{fig:pipeline}, supervised by deformable masks $\mathbf{M}$. Moreover, leveraging K-Planes~\cite{kplanes_2023}, the multiple 4D local deformable space can be further compressed in only $\mathbf{C}_4^2=6$ feature planes. We iteratively sample from coarse to fine within the multi-scale feature plane, retrieving the maximum probability interaction variables $\boldsymbol{\kappa}_u$ and indices $u$ at each scale.

\begin{figure}[htbp]
    % \vspace{-1ex}
    \begin{center}
        \includegraphics[width=0.8\linewidth]{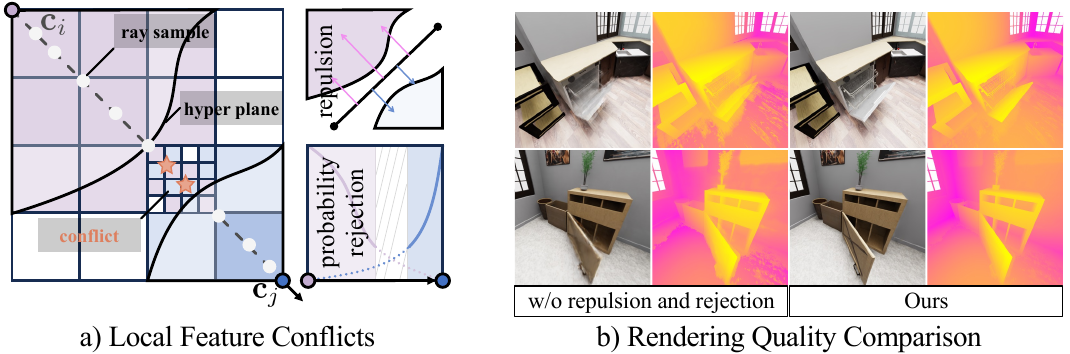}
    \end{center}
    \vspace{-2ex}
    \caption{Illustration of a) boundary sampling conflicts, b) rendering quality comparison.}
    \label{fig:conflicts}
    \vspace{-2ex}
\end{figure}

\textbf{Feature Repulsion and Probability Rejection.}
A latent challenge is that optimizing the interaction probability decoder with varying masks can lead to blurred boundaries in the local deformable field, further causing ray sampling and feature storage conflicts. As illustrated in~\cref{fig:conflicts}(a), consider two adjacent local deformable regions $\mathcal{R}_i$ and $\mathcal{R}_j$, and a point $\mathbf{p}$ in high-dimensional space, suppose $\mathbf{p}$ moves from the cluster center of $\mathcal{R}_i$ towards the cluster center of $\mathcal{R}_j$, then the probability of $\mathbf{p}$ belonging to $\mathcal{R}_i$ gradually decreases, while the probability of $\mathbf{p}$ belonging to $\mathcal{R}_j$ increases. To avoid sampling conflicts and feature oscillations at the boundaries, we introduce a repulsion loss for ray pairs $(\mathbf{r}_i, \mathbf{r}_j)$, and amplify the feature differences between distinct deformable regions, promoting the separation of deformable field:
\begin{equation}
    \resizebox{0.6\linewidth}{!}{
    \begin{math}
    \begin{aligned}
        \mathcal{L}_\text{repuls} = \mathbf{ELU}(K - \left \| (\mathbf{M}_i \odot \mathbf{M}_j)(\mathcal{F}_i - \mathcal{F}_j) \right \|),
    \end{aligned}
    \end{math}
    }
\end{equation}
where $K$ is a constant hyperparameter, $\mathbf{M}_i$ and $\mathbf{M}_j$ are the ground truth mask of rays. $\mathcal{F}_i$ and $\mathcal{F}_j$ are the last-layer features of interaction probability decoder in~\cref{fig:pipeline}. During training, we randomly select ray pairs and apply $\mathcal{L}_\text{repuls}$ to enforce the separation of interactive probability features across local deformable spaces. The initiative is inspired by~\cite{garfield_kim_2024}, which demonstrated the efficacy of repulsive forces in disambiguating 3D segmentation results.

Additionally, the probability rejection is proposed to truncate the low-probability samples if the maximum deformable probability $\mathbf{P}$ at sample $\mathbf{p}$ is smaller than threshold $s$, and selects the background feature directly. The operation is defined as:
\begin{equation}
    \begin{aligned}
        u=\left\{\begin{array}{ll}
                     \argmax_{i} \{\mathbf{P}_i\}, & \text {if}  \quad \mathbf{P}_i \geq s  \\
                     -1,                                                                           & \text { otherwise }
                 \end{array}\right..
    \end{aligned}
\end{equation}
As shown in~\cref{fig:conflicts}(b), the proposed operations help the model achieve higher rendering quality, demonstrating their effectiveness in alleviating boundary sampling conflicts.
\subsection{Interaction-Aware Language Embedding}
\label{sec:language}
Language embedding in interactive scenes is complex and storage-intensive, as 3D distillation faces the dual challenges of high-dimensional optimization and interaction scene variation inconsistency, such as the distinct topological structures of a transformer toy before and after transformation, leading to the failure of SAM~\cite{kirillov2023segany} segmentation or LERF~\cite{lerf_kerr_2023} grounding. As shown in~\cref{fig:pipeline}, leveraging the proposed multi-scale interaction space factorization of~\ref{sec:space_factor}, we efficiently store language features in lightweight planes by indexing them according to maximum probability sampling instead of 3D fields in LERF. For any sampling point $\mathbf{p}$, we project it onto $\mathbf{p}_u =\left [ \mathbf{x} | \boldsymbol{\kappa}(u) \right ]$, retrieving a local language feature group by index $d$, and perform bilinear interpolation using $\boldsymbol{\kappa}_u$ to obtain a language embedding that adapts to interactive variable changes from surrounding clip features. By interpolating language embeddings, our method not only perceives topological structure changes but also achieves a storage complexity of $\mathbf{O}(C \times \alpha \times \mathbf{dim})$, much smaller than language distillation methods like LERF that operate in 3D scenes, where $\mathbf{dim}$ is the dimension of CLIP feature.
\section{Dataset}
\label{sec:dataset}
\begin{figure}[t]
    \vspace{-2ex}
    \begin{center}
        \includegraphics[width=0.98\linewidth]{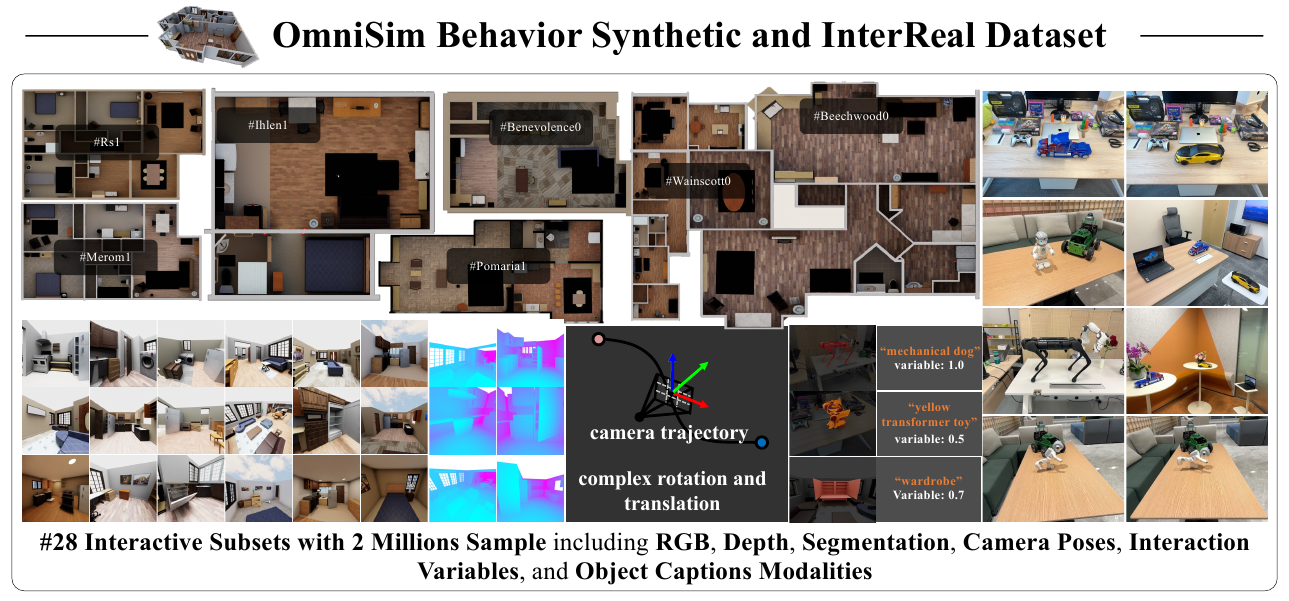}
    \end{center}
    \vspace{-2ex}
    \caption{\textbf{Overview of the \simdata and \realdata datasets}.}
    \label{fig:dataset}
    \vspace{-4ex}
\end{figure}
To our knowledge, existing view synthetic datasets for interactive scene rendering are primarily limited to a few interactive objects~\cite{yunus2024recent,conerf_kania_2022,yu2023cogs,zheng2023editablenerf} due to necessitating a substantial amount of manual annotation of object masks and states, making it impractical to scale up to real scenarios involving multi-object interactions. To bridge this gap, we construct two scene-level, high-quality annotated datasets to advance research progress in reconstructing and understanding interactive scenes: \textbf{\simdata} and \textbf{\realdata}, as shown in~\cref{fig:dataset}. Besides, we use the CoNeRF Synthetic and Controllable~\cite{conerf_kania_2022} dataset for evaluation as well. \textbf{1) \simdata Dataset} is rendered through OmniGibson~\cite{li2024behavior} simulator, leveraging 7 indoor scene models: \texttt{\#rs}, \texttt{\#ihlen}, \texttt{\#beechwod}, \texttt{\#merom}, \texttt{\#pomaria}, \texttt{\#wainscott} and \texttt{\#benevolence}. By varying the rotation vectors of the articulated objects' joints and the camera's trajectory within the scene, we generated 20 high-definition subsets, each consisting of RGBD images, camera trajectory, interactive object masks, and corresponding object state quantities at each time step. \textbf{2) \realdata Dataset} is captured from 8 real Interactable scenes and finely annotated with interaction variables and masks, camera poses encompassing multiple objects, and articulated motion variables. More details can be found in the supplementary.
\section{Experiment}
\label{sec:experiment}
\boldparagraph{Baselines.} We compare~\ours with the existing 3D static rendering methods~\cite{mildenhall2020nerf,instant_mller_2022,3d_kerbl_2023,kerbl3Dgaussians}, 4D deformable methods~\cite{kplanes_2023,park2021hypernerf,park2021nerfies}, and controllable scene reconstruction methods~\cite{conerf_kania_2022,yu2023cogs}. Note that we reimplemented CoGS~\cite{yu2023cogs} based on Deformable Gaussian~\cite{yang2023deformable3dgs} since the official code is unavailable. Additionally, we extended K-Planes~\cite{kplanes_2023} from $\mathbf{C}_4^2$ planes to $\mathbf{C}_{3+\alpha}^2$ planes, denoted as MK-Planes, where $\alpha$ represents the number of interactable objects in dynamic scenes. By leveraging the fact that each instance occupies a distinct region, we further compressed the model, denoted as MK-Planes$^\star$, requiring only $3+3\alpha$ planes.

\boldparagraph{Implementation.} \ours is implemented in Nerfstudio~\cite{nerfstudio_tancik_2023} from scratch. We represent the field as a multi-scale feature plane with resolutions of  $512 \times 256 \times 128$, and feature dimension of 32. The proposal network adopts a coarse-to-fine sampling process, where each sampling step concatenates the position feature and the state quantity as the query for the 4D deformation probability field, which is a 1-layer MLP with 64 neurons and ReLU activation. We use the Adam optimizer with initial learning rates of 0.01 and a cosine decay scheduler with 512 warmup steps for all networks. The model is trained with an NVIDIA A100 GPU for 80k steps on the \simdata dataset and 100k steps on the \realdata dataset, using a batch size of 4096 rays with 64 samples. More implementation can be found in the supplemental materials.

\subsection{View Synthesis Quality Comparison}
\begin{figure}[t]
    \vspace{-2ex}
    \begin{center}
        \includegraphics[width=1\linewidth]{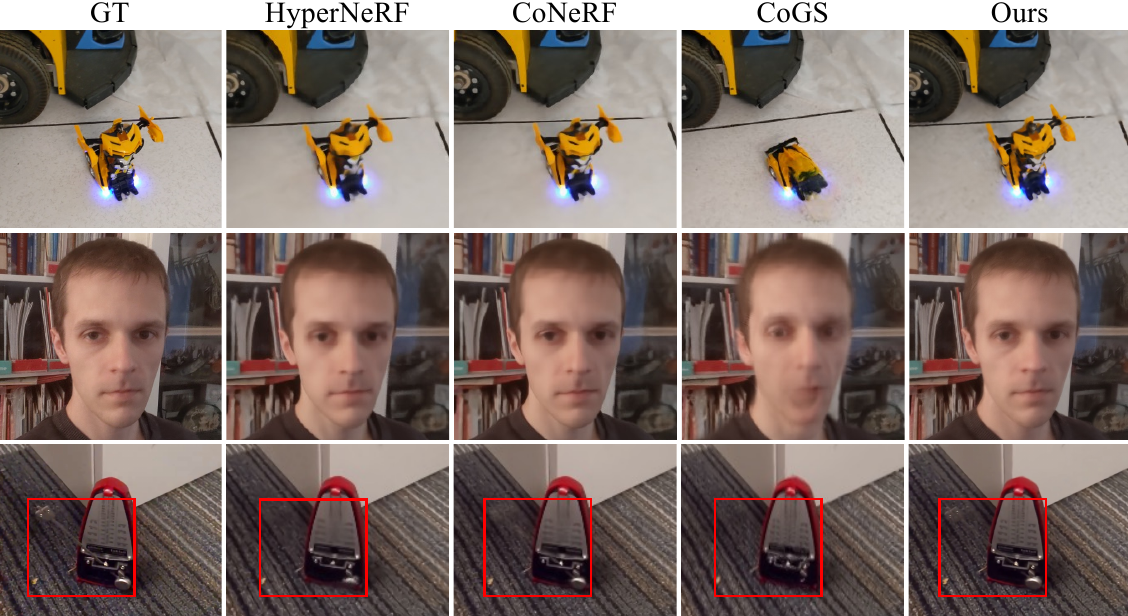}
    \end{center}
    \vspace{-2ex}
    \caption{\textbf{View Synthesis Visualization on CoNeRF Controllable Dataset}. The proposed method achieves higher-quality rendering results compared with the existing methods.}
    \label{fig:render_conerf}
    \vspace{-2ex}
\end{figure}

\textbf{Evaluation on CoNeRF Synthetic and Controllable Datasets.}
We report the quantitative results on CoNeRF Synthetic and Controllable scenes in~\cref{tab:conerf}. \ours outperforms all the existing PSNR, SSIM, and LPIPS metrics methods on CoNeRF Synthetic scenes with a large margin. In particular, \ours achieves 43.349, 0.986, and 0.011 in PSNR, SSIM, and LPIPS, respectively, outperforming the second-best method by 9.894, 0.009, and 0.053. On CoNeRF Controllable, \ours achieves the best PSNR of 32.782 and comparable SSIM and LPIPS to the SOTA methods. According to the novel view synthesis results in~\cref{fig:render_conerf}, \ours achieves more detailed and higher rendering quality, demonstrating the effectiveness of \ours in modeling object-level interactive scenarios. 

\begin{table}[t]
    % \vspace{-2ex}
    \centering
    \setlength{\tabcolsep}{8pt}
    \renewcommand{\arraystretch}{0.60}
    \caption{\textbf{Quantitative results on CoNeRF synthetic and controllable datasets.} \ours achieves the best results in all metrics on synthetic scenes and the best PSNR on the controllable datasets.}
    \resizebox{0.75\columnwidth}{!}{%
        \begin{tabular}{lccccccc}
            \toprule
            \multirow{3}{*}{Method}                      & \multicolumn{3}{c}{CoNeRF Syntetic} &                & \multicolumn{3}{c}{CoNeRF Controllable}                                                          \\ \cmidrule{2-4} \cmidrule{6-8}
                                                         & PSNR$\uparrow$                      & SSIM$\uparrow$ & LPIPS$\downarrow$                       &  & PSNR$\uparrow$ & SSIM$\uparrow$ & LPIPS$\downarrow$ \\
            \midrule
            NeRF~\cite{mildenhall2020nerf}               & 25.299                              & 0.843          & 0.197                                   &  & 28.795         & 0.951          & 0.210             \\
            InstantNGP~\cite{instant_mller_2022}         & 27.057                              & 0.903          & 0.230                                   &  & 26.391         & 0.884          & 0.278             \\
            3DGS~\cite{kerbl3Dgaussians}                 & 32.576                              & \nd 0.977      & 0.077                                   &  & 25.945         & 0.834          & 0.414             \\
            \midrule
            NeRF + Latent~\cite{mildenhall2020nerf}      & 28.447                              & 0.939          & 0.115                                   &  & \nd 32.653     & \nd 0.981      & 0.182             \\
            Nerfies\cite{park2021nerfies}                & ---                                 & ---            & ---                                     &  & 32.274         & \nd 0.981      & 0.180             \\
            HyperNeRF\cite{park2021hypernerf}            & 25.963                              & 0.854          & 0.158                                   &  & 32.520         & \nd 0.981      & 0.169             \\
            K-Planes~\cite{kplanes_2023}                 & 33.301                              & 0.933          & 0.150                                   &  & 31.811         & 0.912          & 0.262             \\
            \midrule
            CoNeRF-$\mathcal{M}$\cite{conerf_kania_2022} & 27.868                              & 0.898          & 0.155                                   &  & 32.061         & 0.979          & \nd 0.167         \\
            CoNeRF\cite{conerf_kania_2022}               & 32.394                              & 0.972          & 0.139                                   &  & 32.342         & \nd 0.981      & 0.168             \\
            CoGS~\cite{yu2023cogs}                       & \nd 33.455                          & 0.960          & \nd 0.064                               &  & 32.601         & \fs 0.983      & \fs 0.164         \\
            \ours (Ours)                                 & \fs 43.349                          & \fs 0.986      & \fs 0.011                               &  & \fs 32.782     & 0.932          & 0.186             \\
            \bottomrule
        \end{tabular}%
    }
    \label{tab:conerf}
    % \vspace{-2.0ex}
\end{table}

\begin{table}[t]
    \vspace{-2.0ex}
    \centering
    \setlength{\tabcolsep}{2pt}
    \renewcommand{\arraystretch}{0.8}
    \caption{\textbf{Quantitative results on \simdata Dataset}. \ours outperforms prior works on most metrics and achieves the best PSNR on the \#challenging subset with a significant margin.}
    \resizebox{1\columnwidth}{!}{
        \begin{tabular}{lccccccccccccccc}
            \toprule
            \multirow{3}{*}{Method}              & \multicolumn{3}{c}{\#Easy Sets} &                & \multicolumn{3}{c}{\#Medium Sets} &  & \multicolumn{3}{c}{\#Challenging Sets} &                & \multicolumn{3}{c}{\#Avg (all 20 Sets)}                                                                                                                   \\
            \cmidrule{2-4} \cmidrule{6-8} \cmidrule{10-12} \cmidrule{14-16}
                                                 & PSNR$\uparrow$                  & SSIM$\uparrow$ & LPIPS$\downarrow$                 &  & PSNR$\uparrow$                         & SSIM$\uparrow$ & LPIPS$\downarrow$                       &  & PSNR$\uparrow$ & SSIM$\uparrow$ & LPIPS$\downarrow$ &  & PSNR$\uparrow$ & SSIM$\uparrow$ & LPIPS$\downarrow$ \\
            \midrule
            NeRF~\cite{mildenhall2020nerf}       & 25.817                          & 0.906          & 0.167                             &  & 25.645                                 & 0.928          & 0.138                                   &  & 26.364         & 0.927          & 0.128             &  & 25.776         & 0.916          & 0.153             \\
            InstantNGP~\cite{instant_mller_2022} & 25.704                          & 0.902          & 0.183                             &  & 25.627                                 & 0.930          & 0.140                                   &  & 26.367         & 0.920          & 0.143             &  & 25.706         & 0.914          & 0.164             \\
            HyperNeRF~\cite{park2021hypernerf}   & 30.708                          & 0.908          & 0.316                             &  & 31.621                                 & 0.936          & 0.265                                   &  & 27.533         & 0.897          & 0.318             &  & 30.748         & 0.917          & 0.299             \\
            K-Planes~\cite{kplanes_2023}         & \nd 32.841                      & 0.952          & \nd 0.093                         &  & 32.548                                 & \nd 0.954      & 0.100                                   &  & 29.833         & 0.937          & 0.118             &  & 32.573         & 0.952          & \nd 0.097         \\
            \midrule
            CoNeRF~\cite{conerf_kania_2022}      & 32.104                          & 0.932          & 0.254                             &  & \nd 33.256                             & 0.951          & 0.207                                   &  & \nd 30.349     & 0.923          & 0.238             &  & 32.477         & 0.939          & 0.234             \\
            MK-Planes$^\star$                    & 31.630                          & 0.948          & 0.098                             &  & 31.880                                 & 0.951          & 0.104                                   &  & 26.565         & 0.887          & 0.218             &  & 31.477         & 0.946          & 0.106             \\
            MK-Planes                            & 31.677                          & 0.948          & 0.098                             &  & 32.165                                 & 0.952          & 0.099                                   &  & 29.254         & 0.933          & 0.119             &  & 31.751         & 0.949          & 0.099             \\
            CoGS~\cite{yu2023cogs}               & 32.315                          & \nd 0.961      & 0.108                             &  & 32.447                                 & \fs 0.965      & \nd 0.086                               &  & 28.701         & \fs 0.970      & \fs 0.073         &  & 32.187         & \fs 0.963      & \nd 0.097         \\
            \ours (Ours)                         & \fs 33.221                      & \fs 0.962      & \fs 0.072                         &  & \fs 33.262                             & \fs 0.965      & \fs 0.072                               &  & \fs 31.645     & \nd 0.948      & \nd 0.093         &  & \fs 33.158     & \nd 0.962      & \fs 0.074         \\
            \bottomrule
        \end{tabular}
    }
    \label{tab:omnisim}
\end{table}

\begin{table}[!htbp]
    \vspace{-2.0ex}
    \centering
    \setlength{\tabcolsep}{2pt}
    \renewcommand{\arraystretch}{0.75}
    \caption{\textbf{Quantitative results on \realdata Dataset}. Our method outperforms others in most settings, with a significant advantage of PSNR, SSIM, and LPIPS on the \#challenging subset.}
    \resizebox{0.85\columnwidth}{!}{
        \begin{tabular}{lccccccccccc}
            \toprule
            \multirow{3}{*}{Method}              & \multicolumn{3}{c}{\#Medium Sets} &                & \multicolumn{3}{c}{\#Challenging Sets} &  & \multicolumn{3}{c}{\#Avg (all 8 Sets)}                                                                                               \\
            \cmidrule{2-4} \cmidrule{6-8} \cmidrule{10-12}
                                                 & PSNR$\uparrow$                    & SSIM$\uparrow$ & LPIPS$\downarrow$                      &  & PSNR$\uparrow$                         & SSIM$\uparrow$ & LPIPS$\downarrow$ &  & PSNR$\uparrow$ & SSIM$\uparrow$ & LPIPS$\downarrow$ \\
            \midrule
            NeRF~\cite{mildenhall2020nerf}       & 20.816                            & 0.682          & 0.190                                  &  & 21.169                                 & 0.728          & 0.337             &  & 20.905         & 0.694          & 0.227             \\
            InstantNGP~\cite{instant_mller_2022} & 21.700                            & 0.776          & 0.215                                  &  & 21.643                                 & 0.745          & 0.338             &  & 21.686         & 0.769          & 0.245             \\
            HyperNeRF~\cite{park2021hypernerf}   & 25.283                            & 0.671          & 0.467                                  &  & 25.261                                 & 0.713          & 0.517             &  & 25.277         & 0.682          & 0.480             \\
            K-Planes~\cite{kplanes_2023}         & 27.999                            & 0.813          & 0.177                                  &  & 26.427                                 & \nd 0.756      & \nd 0.331         &  & 27.606         & 0.799          & 0.215             \\
            CoNeRF~\cite{conerf_kania_2022}      & 27.501                            & 0.745          & 0.367                                  &  & \nd 26.447                             & 0.734          & 0.472             &  & 27.237         & 0.742          & 0.393             \\
            CoGS~\cite{yu2023cogs}               & \nd 30.774                        & \fs 0.913      & \nd 0.100                              &  & \blackx                                & \blackx        & \blackx           &  & \fs 30.774     & \fs 0.913      & \nd 0.100         \\
            \ours (Ours)                         & \fs 30.815                        & \nd 0.911      & \fs 0.066                              &  & \fs 28.436                             & \fs 0.846      & \fs 0.185         &  & \nd 30.220     & \nd 0.895      & \fs 0.096         \\
            \bottomrule
        \end{tabular}
    }
    \label{tab:intereal}
    \vspace{-2.0ex}
\end{table}

\textbf{Evaluation on \simdata Datasets}. \simdata dataset is categorized into 3 interaction level subsets: \#easy, \#medium, and \#challenging, based on the number of interactive objects in each scene. As shown in~\cref{tab:omnisim}, \ours achieves the best PSNR, SSIM, and LPIPS on all interaction level subsets of \simdata, with average PSNR, SSIM, and LPIPS of 33.158, 0.962, and 0.074, respectively. Notably, substantial performance degradation is observed across all methods as the quantity and complexity of interactive objects increase, e.g., CoGS~\cite{yu2023cogs} experiences a 3.641 dB PSNR drop from \#easy to \#challenging. While \ours maintains a relatively stable high performance across all subsets, demonstrating its robustness in modeling complex interactive scenarios.

\textbf{Evaluation on \realdata Datasets}. We divided \realdata dataset into \#medium and \#challenging subsets. In ~\cref{tab:intereal}, CoGS~\cite{yu2023cogs} underperforms compared to \ours on the \#medium subset and fails to converge when faced with long camera trajectories and a large number of interactive objects in the scene (\#challenging), highlighting the limitation of existing methods in modeling real-world interactive scenarios. In contrast, \ours achieves the highest PSNR of 28.436 and the lowest LPIPS of 0.185 on the \#challenging subset, indicating its superiority in modeling real-world large-scale interactive scenarios.

\begin{figure}[!ht]
    % \vspace{-2ex}
    \begin{center}
        \includegraphics[width=1\linewidth]{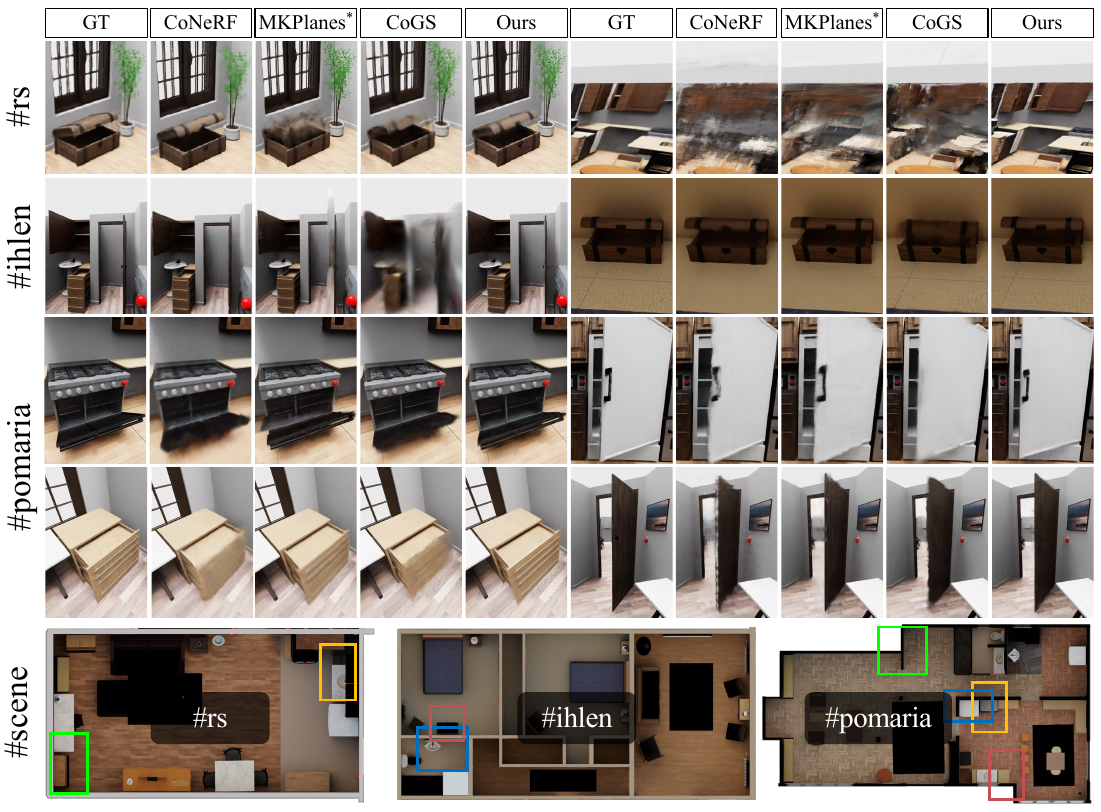}
    \end{center}
    \vspace{-2ex}
    \caption{\textbf{View Synthesis Visualization on \simdata Dataset}. We compare our method with SOTA methods on RGB rendering across three scenes: \#rs, \#ihlen, and \#pomaria. Boxes of different colors represent distinct interactive objects within the scene.}
    \label{fig:render}
    \vspace{-2ex}
\end{figure}

\textbf{View Synthesis Visulization}. \cref{fig:render} presents the novel view synthesis results of \ours and the SOTA methods on \simdata dataset. The results reveal that \ours generates more detailed results than SOTA methods, particularly in complex interactive scenarios. For instance, on the \#pomaria scene featuring an openable dishwasher, CoNeRF~\cite{conerf_kania_2022} fails to capture details, while CoGS~\cite{yu2023cogs} and MK-Planes$^\star$ exhibit residual artifacts. In contrast, our method accurately reconstructs the internal details. Another challenge arises in the \#rs scene, where other methods struggle to reconstruct distant and static objects. In comparison, our method not only overcomes the challenging problem of dramatic topology changes in interactive scenes but also maintains the ability to reconstruct high-quality static scenes.

% \vspace{-1.0ex}
\subsection{Language Grounding Comparison}
\begin{figure}[t]
    % \vspace{-2ex}
    \centering
    \footnotesize
    \begin{minipage}[!b]{0.48\textwidth}
        \centering
        \setlength{\tabcolsep}{3.0pt}
        \renewcommand{\arraystretch}{0.9}
        \begin{tabular}{cclcccc}
            \midrule
            \multicolumn{6}{c}{mIOU $\uparrow$}                                                                                                                         \\
            \midrule
            \multicolumn{3}{c}{setting}                                    & SAM~\cite{kirillov2023segany} & LERF~\cite{lerf_kerr_2023} & Ours                          \\
            \multirow{4}{*}{\rotatebox[origin=l]{90}{\scriptsize\simdata}} &                               & \#easy                     & \nd 61.58 & 23.60 & \fs 86.94 \\
                                                                           &                               & \#medium                   & \nd 55.13 & 19.40 & \fs 86.32 \\
                                                                           &                               & \#challenging              & \nd 63.86 & 19.87 & \fs 90.41 \\
                                                                           &                               & \#avg                      & \nd 59.11 & 21.74 & \fs 86.86 \\
            \midrule
            \multirow{3}{*}{\rotatebox[origin=l]{90}{\scriptsize\realdata}} &                              & \#medium                   & 93.27 & 27.63 & \nd 84.37          \\
                                                                            &                              & \#challenging              & \fs 91.50 & 34.39 & \fs 91.90 \\
                                                                            &                              & \#avg                      & \fs 92.82 & 29.32 & \nd 86.26          \\
            \midrule
        \end{tabular}
    \end{minipage}
    \hfill
    \begin{minipage}[!b]{0.51\textwidth}
        \centering
        \includegraphics[width=\textwidth]{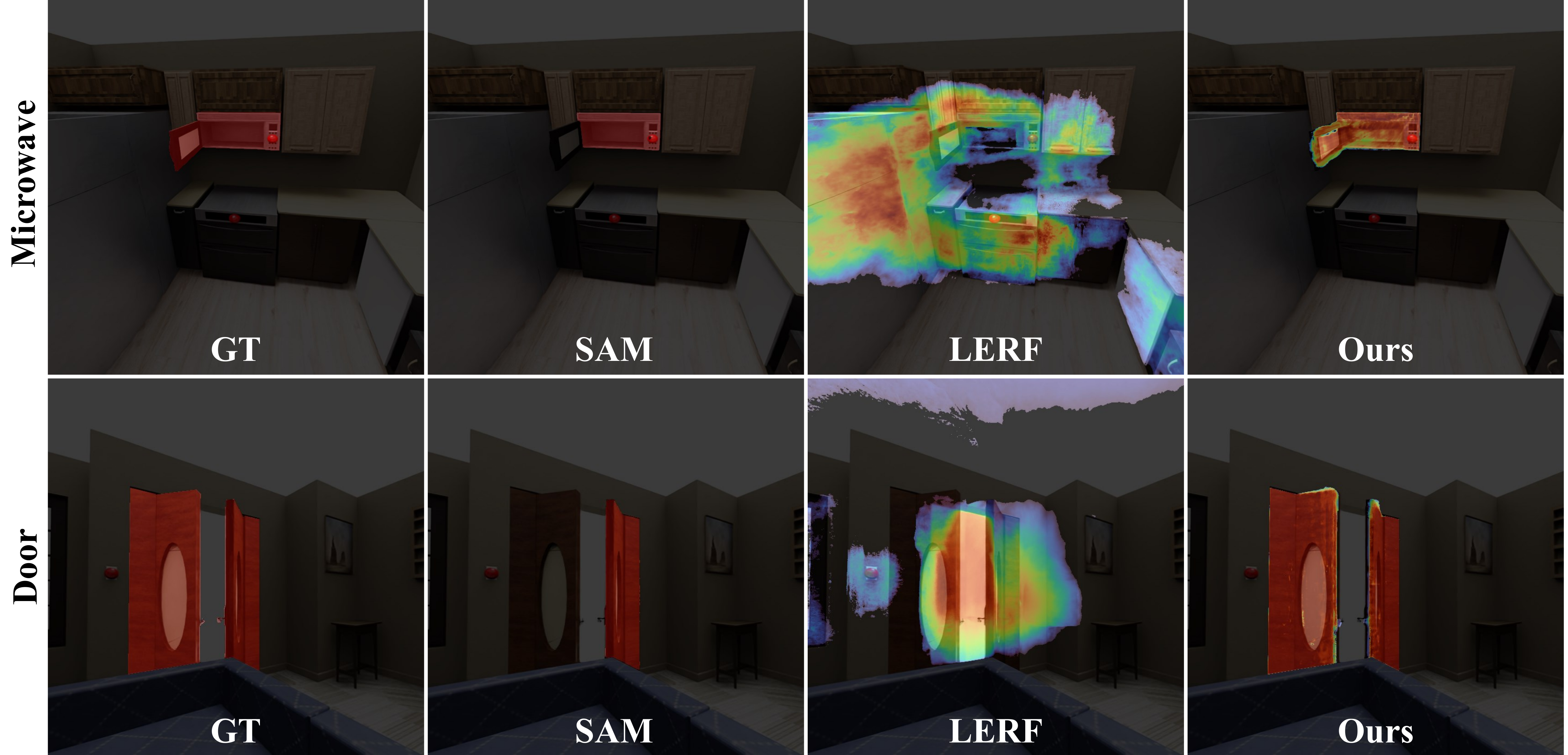}
    \end{minipage}
    \vspace{-1ex}
    \caption{\textbf{Language Grounding Performance on \simdata Dataset} left): Our method gains the highest mIOU score. right): LiveScene's grounding exhibits clearer boundaries than other methods.}
    \label{fig:miou}
    % \vspace{-3ex}
\end{figure}

We assess the language grounding performance on the \simdata dataset using mIOU metric. \cref{fig:miou} suggests that our method obtains the highest mIOU score, with an average of 86.86. In contrast, traditional methods like LERF~\cite{lerf_kerr_2023} encounter difficulties in locating objects precisely, with an average mIOU of 21.74. Meanwhile, 2D methods like SAM~\cite{kirillov2023segany} fail to accurately segment the whole target under specific viewing angles, as objects appear discontinuous in the image. Conversely, our method perceives the completeness of the object and has clear knowledge of its boundaries, demonstrating its advantage in language grounding tasks.

\begin{figure}[h]
    \vspace{-2ex}
    \begin{center}
        \includegraphics[width=1\linewidth]{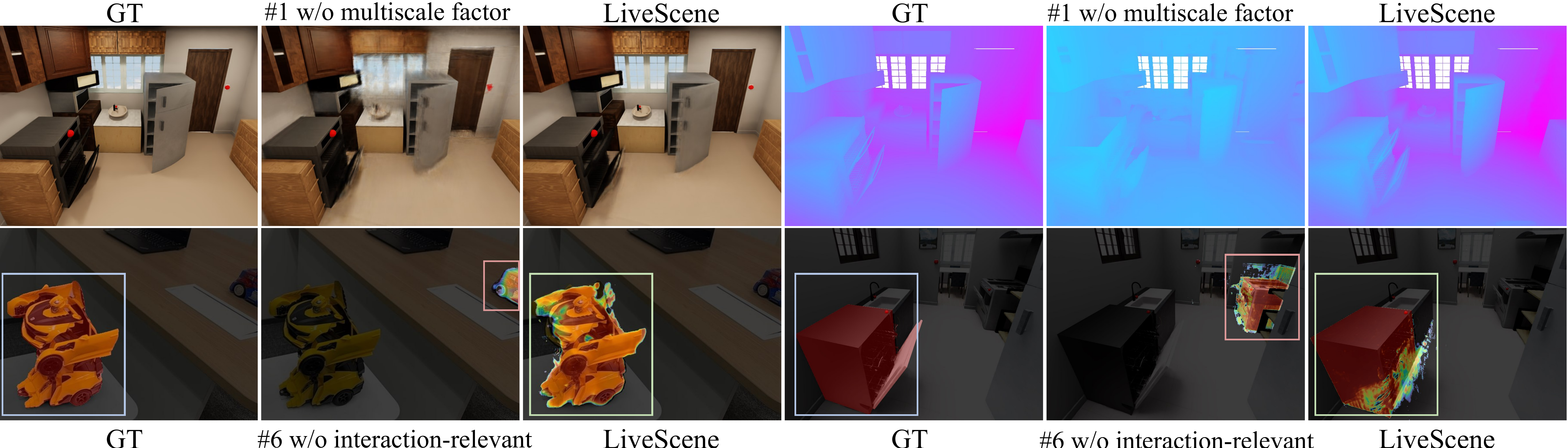}
    \end{center}
    \vspace{-1ex}
    \caption{\textbf{Rendering and Grounding Performance for \#1 and \#6.} above): Multi-scale factorization greatly boosts the performance of RGB rendering and geometry reconstruction. below): Without view consistency, the model struggles when objects have similar appearances.}
    \label{fig:ablation}
    % \vspace{-4ex}
\end{figure}

\begin{figure}[ht]
    \vspace{-2ex}
    \begin{center}
        \includegraphics[width=1\linewidth]{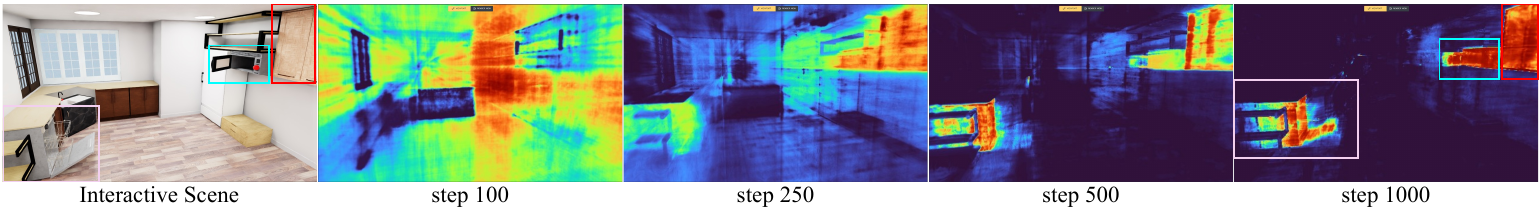}
    \end{center}
    \vspace{-2ex}
    \caption{Learning process of the probability fields from 0 to 1000 training steps. The model progressively converges to the vicinity of the interactive objects, establishing interactive regions.}
    \label{fig:training}
    % \vspace{-2ex}
\end{figure}

% \vspace{-1.5ex}
\subsection{Ablation Study}
In this section, we present ablative studies to investigate the effectiveness of each component in \ours. We selected 1 scene from the \#medium subset, 2 scenes from the \#easy subset of \simdata dataset, and 1 scene each from the \#medium and \#challenging settings for \realdata. Notably, ground-truth state quantities are only available in \simdata, not in \realdata. Therefore, we use GT quantities on \simdata and introduce a learnable variable on \realdata to infer state changes. \cref{fig:ablation} reports the rendering quality and grounding performance for \#1 and \#6.

\boldparagraph{Effectiveness of components.} As illustrated in~\cref{tab:ablation}, the multi-scale factorization significantly improves the rendering performance on both datasets, with PSNR on \simdata increasing from 31.74 to 35.094, shown in \#1. Introducing learnable variables for each frame (\#2) yields corresponding improvements on \realdata dataset since this latent code can perceive the change in object states. The feature repulsion loss and probability rejection (\#3 and \#4) together make rendering quality better in \realdata as well as in \simdata dataset. As for grounding, \#5 shows that rendering embeddings along a ray~\cite{decomposing_kobayashi_2022,inplace_zhi_2021} struggles to locate objects precisely. Ensuring view consistency further boosts grounding performance on \simdata, as demonstrated in \#6.

\boldparagraph{Probability field training.} We provide additional experiments in~\cref{fig:training} of the disjoint regions to illustrate the learning process of the probability field from 0 to 1000 training steps. The results demonstrate a clear trend that, as training advances, the proposed method can progressively converge to the vicinity of the interactive objects, thereby establishing interactive regions. With the establishment of the probability field, the model can focus on different interactive objects and guide the sampling process by maximizing probability, thereby achieving disentanglement of interactive scenes.

\begin{table}[h]
    \centering
    \vspace{-2ex}
    \setlength{\tabcolsep}{8pt}
    \renewcommand{\arraystretch}{0.8}
    \caption{\textbf{Ablation Study} on the subset of \realdata and \simdata Datasets.}
    \resizebox{0.98\columnwidth}{!}{
        \begin{threeparttable}
            \begin{tabular}{clcccccccccccccc}
                \toprule
                \multirow{2}{*}{\#exp} & \multicolumn{6}{c}{\#settings~(rendering)} &                                      & \multicolumn{3}{c}{\realdata}       &              & \multicolumn{4}{c}{\simdata}                                                                                                                                                                                            \\ \cmidrule{2-7} \cmidrule{9-11} \cmidrule{13-16}
                                       & \Romannum{1}                               & \Romannum{2}                         & \Romannum{3}                        & \Romannum{4} & \Romannum{5}                        & \Romannum{6} &  & PSNR$\uparrow$            & SSIM$\uparrow$          & LPIPS$\downarrow$         &  & PSNR$\uparrow$ & SSIM$\uparrow$ & LPIPS$\downarrow$ & Depth L1$\downarrow$ \\
                \midrule
                \#0                    &                                            &                                      &                                     &              &                                     &              &  & \textcolor{gray}{25.329}  & \textcolor{gray}{0.731} & \textcolor{gray}{0.329}   &  & 31.740          & 0.938          & 0.118             & 0.238                \\
                \#1                    & \bluecheck                                 &                                      &                                     &              &                                     &              &  & 28.289                    & 0.819                   & 0.226                     &  & 35.094         & 0.969          & 0.059             & 0.086                \\
                % \#2                    & \bluecheck                                 & \bluecheck                           &                                     &              &                                     &              &  & 29.577                    & 0.865                   & 0.162                     &  & \textcolor{gray}{35.826} & \textcolor{gray}{0.973} & \textcolor{gray}{0.053} & 0.096                \\
                \#2                    & \bluecheck                                 & \bluecheck                           &                                     &              &                                     &              &  & 29.577                    & 0.865                   & 0.162                     &  & ---            & ---            & ---               & ---                  \\
                \#3                    & \bluecheck                                 & \bluecheck\textcolor{mblue}{$^\ast$} &                                     & \bluecheck   &                                     &              &  & 29.959                    & 0.883                   & 0.131                     &  & 34.989         & 0.969          & 0.059             & 0.085                \\
                \#4                    & \bluecheck                                 & \bluecheck\textcolor{mblue}{$^\ast$} & \bluecheck                          &              &                                     &              &  & 30.123                    & 0.884                   & 0.132                     &  & 34.977         & 0.967          & 0.061             & 0.086                \\
                \ours                  & \bluecheck                                 & \bluecheck\textcolor{mblue}{$^\ast$} & \bluecheck                          & \bluecheck   &                                     &              &  & 30.591                    & 0.896                   & 0.115                     &  & 35.254         & 0.971          & 0.057             & 0.042                \\
                                       & \multicolumn{6}{c}{\#settings~(grounding)} &                                      & \multicolumn{3}{c}{mIOU $\uparrow$} &              & \multicolumn{4}{c}{mIOU $\uparrow$}                                                                                                                                                                                     \\ \cmidrule{2-7} \cmidrule{9-11} \cmidrule{13-16}
                \#5                    & \bluecheck                                 & \bluecheck\textcolor{mblue}{$^\ast$} & \bluecheck                          & \bluecheck   &                                     & \bluecheck   &  & \multicolumn{3}{c}{30.40} &                         & \multicolumn{4}{c}{32.87}                                                                                 \\
                \#6                    & \bluecheck                                 & \bluecheck\textcolor{mblue}{$^\ast$} & \bluecheck                          & \bluecheck   & \bluecheck                          &              &  & \multicolumn{3}{c}{93.10} &                         & \multicolumn{4}{c}{71.64}                                                                                 \\
                \ours                  & \bluecheck                                 & \bluecheck\textcolor{mblue}{$^\ast$} & \bluecheck                          & \bluecheck   & \bluecheck                          & \bluecheck   &  & \multicolumn{3}{c}{93.02} &                         & \multicolumn{4}{c}{78.52}                                                                                 \\
                \bottomrule
            \end{tabular}
            \begin{tablenotes}
                \footnotesize
                \item \Romannum{1}: multi-scale factorization, \quad \Romannum{2}: learnable variable, \quad \Romannum{3}: feature repulsion $\mathcal{L}_\text{repuls}$, \quad \Romannum{4}: probability rejection, \quad \Romannum{5}: maximum probability embeds retrieval, \quad \Romannum{6}: interaction-aware language embedding. \quad \bluecheck\textcolor{mblue}{$^\ast$} denotes enable \Romannum{2} for \realdata but disable for \simdata.
            \end{tablenotes}
        \end{threeparttable}
    }
    \label{tab:ablation}
    \vspace{-2ex}
\end{table}

\section{Conclusion and Limitation}
\label{sec:conclusion}
We present \ours, the first language-embedded interactive neural radiance field for complex scenes with multiple interactive objects. A parameter-efficient factorization technique is proposed to decompose interactive spaces into local deformable fields to model individual interactive objects. Moreover, we introduce a novel interaction-aware language embedding mechanism that effectively localizes and controls interactive objects using natural language. Finally, We construct two challenging datasets that contain multiple interactive objects in complex scenes and evaluate the effectiveness and robustness of \ours.

\boldparagraph{Limitations:} The control ability of \ours is limited by label density. Additionally, our natural language control is currently restricted to closed vocabulary, and it is inherently tied to the capabilities of the underlying foundation model, such as OpenCLIP. In future work, we plan to extend our method to enable open-vocabulary grounding and control, increasing the model's flexibility and range of applications.

{\small

\boldparagraph{Acknowledgements.}
This work is partially supported by the Shanghai AI Laboratory, National Key R\&D Program of China (JF-P23KK00072-3-DF), the National Natural Science Foundation of China (62376222), and Young Elite Scientists Sponsorship Program by CAST (2023QNRC001).
\vspace{-6ex}
}
\maketitlesupplementary
% \clearpage
% % \setcounter{page}{1}
% \appendix
% \section{Appendix / supplemental material}
% Optionally include supplemental material (complete proofs, additional experiments and plots) in appendix. All such materials \textbf{SHOULD be included in the main submission.}
% \maketitlesupplementary

\begin{abstract}
    \noindent This supplementary material accompanies the main paper by providing more details for reproducibility as well as additional evaluations and qualitative results to to verify the effectiveness and robustness of \ours:\\
    \noindent $\triangleright$ \textbf{\cref{sec:supp_dataset}}: Configurations of \simdata and \realdata dataset, including scene assets, interaction variables generation, mask and prompts annotation, and dataset visualization. \\
    \noindent $\triangleright$ \textbf{\cref{sec:demo}}: Video demonstration and anonymous link: \href{https://livescenes.github.io}{https://livescenes. github.io}. \\
    \noindent $\triangleright$ \textbf{\cref{sec:impl_detail}}: Additional implementation details. \\
    \noindent $\triangleright$ \textbf{\cref{sec:add_exp}}: Additional experimental results, including more ablation studies, detailed view synthesis quality comparison, interactive scenes geometry comparison and language grounding comparison.
\end{abstract}
\section{Configurations of \simdata and \realdata datasets}
\label{sec:supp_dataset}
\begin{figure}[htbp]
    \vspace{-1ex}
    \begin{center}
        \includegraphics[width=1\linewidth]{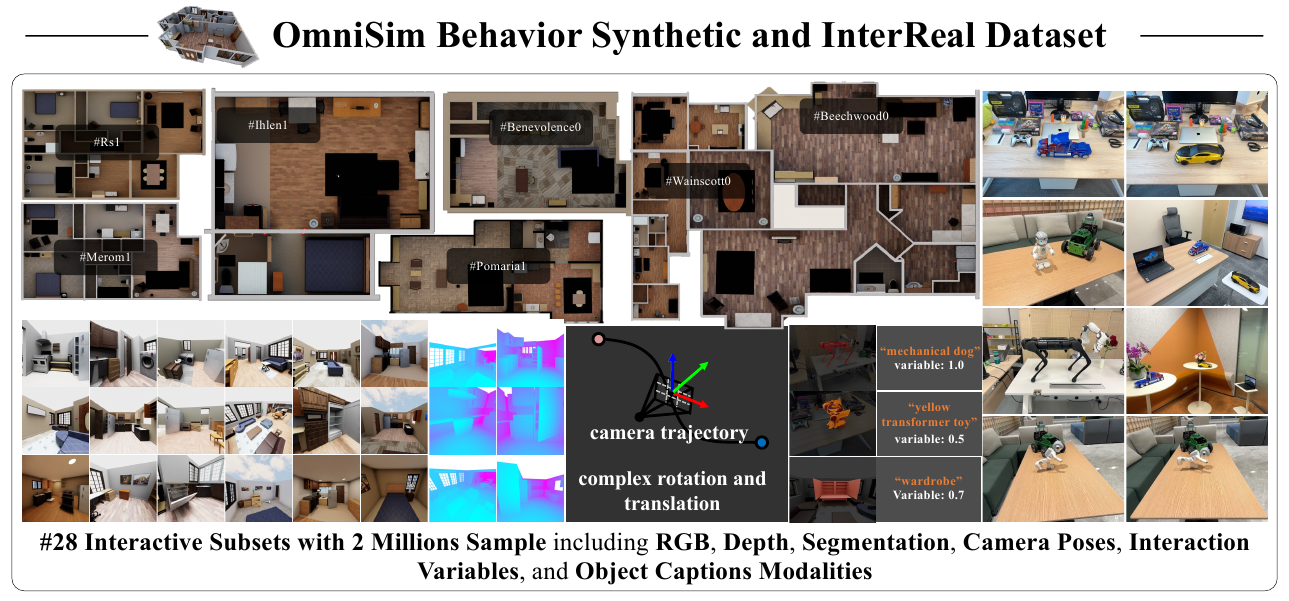}
    \end{center}
    \vspace{-1ex}
    \caption{Illustration of the proposed Omniverse behavior synthetic (OminiSim) and Real captured interactive (InterReal) dataset. These datasets are captured in an OmniGibson simulator or real scene and carefully annotated, providing \#28 interactive subsets with 2 Million samples, including RGB, depth, segmentation, camera trajectory, interaction variables, and object captions modalities.}
    \label{fig:supp_dataset}
    \vspace{-2ex}
\end{figure}

\boldparagraph{Scene Assets and Generation Pipeline for \simdata.}
We generate the synthetic dataset using the OmniGibson simulator. The dataset consists of 20 interactive scenes from 7 scene models: \texttt{\#rs}, \texttt{\#ihlen}, \texttt{\#beechwood}, \texttt{\#merom}, \texttt{\#pomaria}, \texttt{\#wainscott}, and \texttt{\#benevolence}. The scenes feature various interactive objects, including cabinets, refrigerators, doors, drawers, and more, each with different hinge joints.

We configure the simulator camera with an intrinsic parameter set of focal length 8, aperture 20, and a resolution of 1024 × 1024. By varying the rotation vectors for each joint of the articulated objects, we can observe different motion states of various objects. We generated 20 high-definition subsets, each consisting of RGB images, depth, camera trajectory, interactive object masks, and corresponding object state quantities relative to their "closed" state at each time step, from multiple camera trajectories and viewpoints.

The data is obtained through the following steps: 1) The scene model is loaded, and the respective objects are selected, with motion trajectories set for each joint. 2) Keyframes are set for camera movement in the scene, and smooth trajectories are obtained through interpolation. 3) The simulator is then initiated, and the information captured by the camera at each moment is recorded.

\boldparagraph{Scene Assets and Generation Pipeline for \realdata.}
\realdata is primarily captured using the Polycam app on an Apple iPhone 15 Pro. We selected 8 everyday scenes and placed various interactive objects within each scene, including transformers, laptops, microwaves, and more. We recorded 8 videos, each at a frame rate of 5FPS, capturing 700 to 1000 frames per video.

The dataset was processed via the following steps: 1) manual object movement and keyframe capture, 2) OBJ file export and pose optimization using Polycam, 3) conversion to a dataset containing RGB images and transformation matrices using Nerfstudio~\cite{nerfstudio_tancik_2023}, and 4) mask generation for each object in each scene using SAM~\cite{kirillov2023segany} and corresponding prompts and state quantity labeling for certain keyframes.

\boldparagraph{Statistic of \simdata and \realdata Datasets.}
The detailed statistics of the \simdata and \realdata datasets are shown in~\cref{tab:statistic}. The \simdata dataset consists of 20 interactive scenes, each with 2 to 6 objects, and the \realdata dataset contains 8 real-world scenes, each with 1 to 3 objects. The datasets include RGB, depth, pose, mask, and text prompts modalities, providing a total of 2 million samples for training and evaluation. The objects in the datasets include cabinets, refrigerators, doors, drawers, transformers, laptops, microwaves, and more, with various interactive states and captions.

\begin{table}[ht]
    \centering
    \setlength{\tabcolsep}{8pt}
    \renewcommand{\arraystretch}{1.0}
    \caption{\textbf{Statistic of \simdata and \realdata Datasets}.}
    \resizebox{1\columnwidth}{!}{
        \begin{tabular}{clccccccccc}
            \toprule
            \multicolumn{2}{c}{datasets}                         & \#objects                     & \#frame & \#key frame value & rgb  & depth       & pose        & mask        & text prompts                                                           \\
            \midrule
            \multirow{20}{*}{\rotatebox[origin=l]{90}{\simdata}} & \#seq001\_Rs\_int             & 4       & 770               & 770  & \blackcheck & \blackcheck & \blackcheck & \blackcheck  & fridge, microwave, oven, top cabinet                    \\
                                                                 & \#seq002\_Rs\_int             & 4       & 2190              & 2190 & \blackcheck & \blackcheck & \blackcheck & \blackcheck  & fridge, microwave, oven, top cabinet                    \\
                                                                 & \#seq003\_Ihlen\_1\_int       & 3       & 1610              & 1610 & \blackcheck & \blackcheck & \blackcheck & \blackcheck  & bottom cabinet, dishwasher, top cabinet                 \\
                                                                 & \#seq004\_Ihlen\_1\_int       & 2       & 1630              & 1630 & \blackcheck & \blackcheck & \blackcheck & \blackcheck  & bottom cabinet, cedar chest                             \\
                                                                 & \#seq005\_Beechwood\_0\_int   & 2       & 1370              & 1370 & \blackcheck & \blackcheck & \blackcheck & \blackcheck  & bottom cabinet, door                                    \\
                                                                 & \#seq006\_Beechwood\_0\_int   & 2       & 1610              & 1610 & \blackcheck & \blackcheck & \blackcheck & \blackcheck  & dishwasher, microwave                                   \\
                                                                 & \#seq007\_Beechwood\_0\_int   & 3       & 1450              & 1450 & \blackcheck & \blackcheck & \blackcheck & \blackcheck  & bottom cabinet, door, top cabinet                       \\
                                                                 & \#seq008\_Benevolence\_1\_int & 4       & 1830              & 1830 & \blackcheck & \blackcheck & \blackcheck & \blackcheck  & door, fridge, microwave, top cabinet                    \\
                                                                 & \#seq009\_Benevolence\_1\_int & 2       & 1690              & 1690 & \blackcheck & \blackcheck & \blackcheck & \blackcheck  & cedar chest, door                                       \\
                                                                 & \#seq010\_Merom\_1\_int       & 3       & 1930              & 1930 & \blackcheck & \blackcheck & \blackcheck & \blackcheck  & dishwasher, fridge, microwave, top cabinet              \\
                                                                 & \#seq011\_Merom\_1\_int       & 3       & 1690              & 1690 & \blackcheck & \blackcheck & \blackcheck & \blackcheck  & bottom cabinet, top cabinet, door                       \\
                                                                 & \#seq012\_Pomaria\_1\_int     & 2       & 970               & 970  & \blackcheck & \blackcheck & \blackcheck & \blackcheck  & bottom cabinet, fridge                                  \\
                                                                 & \#seq013\_Pomaria\_1\_int     & 3       & 770               & 770  & \blackcheck & \blackcheck & \blackcheck & \blackcheck  & bottom cabinet, fridge                                  \\
                                                                 & \#seq014\_Wainscott\_0\_int   & 2       & 1850              & 1850 & \blackcheck & \blackcheck & \blackcheck & \blackcheck  & bottom cabinet, cedar chest                             \\
                                                                 & \#seq015\_Wainscott\_0\_int   & 2       & 1350              & 1350 & \blackcheck & \blackcheck & \blackcheck & \blackcheck  & bottom cabinet, door                                    \\
                                                                 & \#seq016\_Wainscott\_0\_int   & 2       & 1170              & 1170 & \blackcheck & \blackcheck & \blackcheck & \blackcheck  & fridge, stove                                           \\
                                                                 & \#seq017\_Benevolence\_1\_int & 6       & 4590              & 4590 & \blackcheck & \blackcheck & \blackcheck & \blackcheck  & cedar chest, door, door, fridge, microwave, top cabinet \\
                                                                 & \#seq018\_Benevolence\_1\_int & 2       & 2050              & 2050 & \blackcheck & \blackcheck & \blackcheck & \blackcheck  & door, top cabinet                                       \\
                                                                 & \#seq019\_Rs\_int             & 2       & 1130              & 1130 & \blackcheck & \blackcheck & \blackcheck & \blackcheck  & fridge, top cabinet                                     \\
                                                                 & \#seq020\_Merom\_1\_int       & 2       & 1990              & 1990 & \blackcheck & \blackcheck & \blackcheck & \blackcheck  & bottom cabinet, door                                    \\
                                                                 & \#demo                        & 5       & 6267              & 6267 & \blackcheck & \blackcheck & \blackcheck & \blackcheck  & cedar chest, door, fridge, oven, top cabinet            \\
                                                                 & \#demo001                     & 4       & 2040              & 2040 & \blackcheck & \blackcheck & \blackcheck & \blackcheck  & fridge, microwave, oven, top cabinet                    \\
                                                                 & \#demo002                     & 3       & 2395              & 2395 & \blackcheck & \blackcheck & \blackcheck & \blackcheck  & dishwasher, microwave, top cabinet                      \\
                                                                 & \#demo003                     & 3       & 2480              & 2480 & \blackcheck & \blackcheck & \blackcheck & \blackcheck  & dishwasher, oven, top cabinet                           \\
                                                                 & \#demo004                     & 3       & 2280              & 2280 & \blackcheck & \blackcheck & \blackcheck & \blackcheck  & dishwasher, fridge, stove                               \\
                                                                 & \#demo005                     & 3       & 1670              & 1670 & \blackcheck & \blackcheck & \blackcheck & \blackcheck  & bottom cabinet, cedar chest, door                       \\
            \midrule
            \multirow{8}{*}{\rotatebox[origin=l]{90}{\realdata}} & \#seq001\_transformer         & 1       & 329               & 38   & \blackcheck & \blackx     & \blackcheck & \blackcheck  & yellow toy car                                          \\
                                                                 & \#seq002\_transformer         & 1       & 329               & 43   & \blackcheck & \blackx     & \blackcheck & \blackcheck  & blue toy car                                            \\
                                                                 & \#seq003\_door                & 1       & 355               & 31   & \blackcheck & \blackx     & \blackcheck & \blackcheck  & door                                                    \\
                                                                 & \#seq004\_dog                 & 1       & 213               & 41   & \blackcheck & \blackx     & \blackcheck & \blackcheck  & black mechanical dog                                    \\
                                                                 & \#seq005\_sit                 & 1       & 913               & 25   & \blackcheck & \blackx     & \blackcheck & \blackcheck  & small white humanoid                                    \\
                                                                 & \#seq006\_stand               & 1       & 899               & 33   & \blackcheck & \blackx     & \blackcheck & \blackcheck  & small white humanoid                                    \\
                                                                 & \#seq007\_flower              & 3       & 620               & 153  & \blackcheck & \blackx     & \blackcheck & \blackcheck  & blue toy car, yellow toy car, black laptop              \\
                                                                 & \#seq008\_office              & 4       & 1087              & 658  & \blackcheck & \blackx     & \blackcheck & \blackcheck  & blue toy car, yellow toy car, black laptop, microwave   \\
            \bottomrule
        \end{tabular}
    }
    \label{tab:statistic}
\end{table}

\section{Videos Demonstration and Anonymous Link}
\label{sec:demo}
We provide a video of our proposed method \ours along with this document to demonstrate the interactive scene reconstruction and multimodal control capabilities. Please refer to the anonymous link: \href{https://livescenes.github.io}{https://livescenes.github.io} for more information.
\section{Additional implementation details}
\label{sec:impl_detail}
\boldparagraph{Loss Functions.} In this section, we provide detailed descriptions of the loss functions used in \ours:
\begin{equation}
    \begin{aligned}
        \mathcal{L}_{total} = \mathcal{L}_\text{MSE} + \lambda_1 \mathcal{L}_\text{focus} + \lambda_2 \mathcal{L}_\text{repuls} + \lambda_3 \mathcal{L}_\text{var} + \lambda_4 \mathcal{L}_\text{lang} + \lambda_5 \mathcal{L}_{\text {smooth}},
    \end{aligned}
\end{equation}

\noindent \textbf{Rendering Loss.} We use the standard NeRF rendering loss, which is the sum of the mean squared error (MSE) between the rendered color and the ground truth color, and the MSE between the rendered depth and the ground truth depth. The loss is computed for each pixel in the image and averaged over the entire image:
\begin{equation}
    \begin{aligned}
        \mathcal{L}_\text{MSE} = \frac{1}{N} \sum_{i=1}^{N} \left \| \mathbf{C}_i - \tilde{\mathbf{C}}_i \right \|^2,
    \end{aligned}
\end{equation}
where $\mathbf{C}_i$ and $\tilde{\mathbf{C}}_i$ are the rendered and ground truth RGB values, respectively, and $N$ is the number of pixels in the image.

\noindent \textbf{Focal Loss.} Due to the predominance of background regions in the images, we employ focal loss to enhance the model's focus on the relatively smaller interactive mask regions:
\begin{equation}
    \begin{aligned}
        \mathcal{L}_\text{focus} = \beta \cdot \left(1 - e^{\sum_{i=1}^{\alpha} \mathbf{M}_i \log(\hat{\mathbf{P}}_i)}\right)^{\gamma} \cdot \left(- \sum_{i=1}^{\alpha} \mathbf{M}_i \log(\hat{\mathbf{P}}_i)\right),
    \end{aligned}
\end{equation}
where $\mathbf{M}$ is the ground truth mask label, $\hat{\mathbf{P}}$ is the probability map rendering from the interactive probability field, $\beta$ is the balancing factor, and $\gamma$ is the focusing parameter. In our experiments, we set $\alpha = 0.5$ and $\gamma = 1.5$.

\noindent \textbf{Repulsion Loss.} To avoid sampling conflicts and feature oscillations at the boundaries, we introduce a repulsion loss to amplify the feature differences between distinct deformable scenes, thereby promoting the separation of deformable field:
\begin{equation}
    \resizebox{0.6\linewidth}{!}{
    \begin{math}
    \begin{aligned}
        \mathcal{L}_\text{repuls} = \mathbf{ELU}(K - \left \| (\mathbf{M}_i \odot \mathbf{M}_j)(\mathcal{F}_i - \mathcal{F}_j) \right \|),
    \end{aligned}
    \end{math}
    }
\end{equation}
where $\mathbf{M}_i$ and $\mathbf{M}_j$ are the ground truth mask of rays, and $\mathcal{F}_i$ and $\mathcal{F}_j$ are the last-layer features of interaction probability decoder in~\cref{fig:pipeline}. $K$ is the constant hyperparameters. In training iteration, we randomly select ray pairs and apply $\mathcal{L}_\text{repuls}$ to enforce the separation of interactive probability features across local deformable spaces. Our approach draws inspiration from~\cite{garfield_kim_2024}, which has shown the effectiveness of repulsive forces in resolving ambiguities in 3D segmentation.

\noindent \textbf{Interaction Variable MSE.} We follow the value MSE in~\cite{conerf_kania_2022} and use the standard MSE loss to supervise the interaction values training:
\begin{equation}
    % \resizebox{0.6\linewidth}{!}{
    % \begin{math}
    \begin{aligned}
        \mathcal{L}_\text{var} = \frac{1}{N} \sum_{i=1}^{N} \left \| \boldsymbol{\kappa}_i - \tilde{\boldsymbol{\kappa}}_i \right \|^2,
    \end{aligned}
    % \end{math}
    % }
\end{equation}
where $\boldsymbol{\kappa}_i$ and $\tilde{\boldsymbol{\kappa}}$ are the predicted and ground truth interaction variables, respectively, and $N$ is the number of ray samples of a batch. Note that we only apply $\mathcal{L}_\text{var}$ to the \realdata dataset and use learnable variables as inputs to the model due to the lack of dense ground truth interaction variables. In \simdata, we directly use the ground truth interaction variables as inputs provided by the simulator to achieve precise control.

\noindent \textbf{Language Embedding L2 Loss.} In \ours implementation, we use the huber loss to supervise the language embedding training. But we do not distill the language embedding in the 3D language field but we store the language embedding in the proposed interaction-aware language feature plane. The loss is defined as:
\begin{equation}
    \begin{aligned}
        \mathcal{L}_\text{lang}(\phi(\mathbf{p}), \tilde \phi(\mathbf{p}))=\left\{\begin{array}{ll}
                                                                                      \frac{1}{2}(\phi(\mathbf{p})-\phi(\mathbf{p}))^{2}                              & \text { if }|\phi(\mathbf{p})-\tilde\phi(\mathbf{p})| \leq \delta \\
                                                                                      \delta\left(|\phi(\mathbf{p})-\tilde\phi(\mathbf{p})|-\frac{1}{2} \delta\right) & \text { otherwise }
                                                                                  \end{array}\right.,
    \end{aligned}
\end{equation}
where $\phi(\mathbf{p})$ and $\tilde \phi(\mathbf{p})$ are the predicted and ground truth language embeddings, respectively, and $\delta$ is the threshold. In our experiments, we set $\delta = 1.0$.

\noindent \textbf{Smoothness Loss.} Inspired by K-Planes~\cite{kplanes_2023}, we use 1D Laplacian (second derivative) filter to smooth the local deformable field feature plane, which helps to reduce the noise in the deformable field and alleviate feature oscillations and sampling conflicts at the sampling boundary:
\begin{equation}
    \begin{aligned}
        \mathcal{L}_{\text {smooth}}(\mathbf{p})=\frac{1}{|L| n^{2}} \sum_{l, i, k}\left\|\mathbf{p}_{l}^{i, k-1}-2 \mathbf{p}_{l}^{i, k}+\mathbf{p}_{l}^{i, k+1}\right\|_{2}^{2},
    \end{aligned}
\end{equation}
where $i$ and $k$ are indices on the plane resolution $n$, and $l$ is the feature planes index.

\boldparagraph{Probability Rejection Operation.}
Additionally, a probability rejection operation is proposed to truncate the low-probability samples if the deformable probability at $\mathbf{p}$ is smaller than threshold $s$. The probability rejection is proposed to truncate the low-probability samples if the maximum deformable probability $\mathbf{P}$ at $\mathbf{p}$ is smaller than threshold $s$ and selects the background feature directly. The operation is defined as:
\begin{equation}
    \begin{aligned}
        u=\left\{\begin{array}{ll}
                     \argmax_{i} \{\mathbf{P}_i\}, & \text {if}  \quad \mathbf{P}_i \geq s  \\
                     -1                                                                           & \text { otherwise }
                 \end{array}\right..
    \end{aligned}
\end{equation}
\boldparagraph{Implementation Details.} \ours is implemented in Nerfstudio~\cite{nerfstudio_tancik_2023} from scratch. We represent the field as a multi-scale feature plane with resolutions of  $512 \times 256 \times 128$, and feature dimension of 32. The proposal network adopts a coarse-to-fine sampling process, where each sampling step concatenates the position feature and the state quantity as the query for the 4D deformation mask field, which is a 1-layer MLP with 64 neurons and ReLU activation. For \realdata, we introduce additional learnable variables bound to each frame to capture changes in object states within the scene. These variables are represented by a plane with a resolution typically half the frame number, with a feature dimension of 4 for most scenes. For all experiments, we use the Adam optimizer with initial learning rates of 0.01 and a cosine decay scheduler with 512 warmup steps for all networks. We set loss weights as follows: $\lambda_1=1e-3, \lambda_2=1e-2, \lambda_3=1e-3, \lambda_4=1.0, \lambda_5=1e-3$. The model is trained for 80k steps on the \simdata dataset and 100k steps on the \realdata dataset, using a batch size of 4096 rays with 64 samples each. We run the model on an NVIDIA A100 GPU, requiring approximately 4 hours and 40GB of memory.

\section{Additional Experimental Results}
\label{sec:add_exp}

\boldparagraph{Model Parameter Efficiency Comparison.}
We compare the number of parameters of \ours with other methods in~\cref{tab:parameters}, varying the number of interactive objects in the scene. The results show that \ours has a constant number of parameters, regardless of the number of interactive objects, making it more efficient than other methods. In contrast, CoGS~\cite{yu2023cogs} has a higher base number of parameters and a linear increase with the number of interactive objects. MK-Planes~\cite{kplanes_2023} exhibits a quadratic increase in parameters with the number of interactive objects. Although NeRF\cite{mildenhall2020nerf} and InstantNGP~\cite{instant_mller_2022} have a low number of parameters, they are limited to 3D static scene reconstruction.

\begin{table}[h]
    % \vspace{-2.5ex}
    \centering
    \setlength{\tabcolsep}{8pt}
    \renewcommand{\arraystretch}{1}
    \caption{\textbf{Model Parameters vs Object Quantity}.}
    \resizebox{0.75\columnwidth}{!}{%
        \begin{tabular}{lccccccc}
            \toprule
            \multirow{3}{*}{Method}              & \multicolumn{6}{c}{\# interactive objects} & \multirow{3}{*}{Trend}                                             \\
            \cmidrule{2-7}
                                                 & 1                                          & 2                      & 3     & 4     & 5     & 6                 \\
            \midrule
            NeRF~\cite{mildenhall2020nerf}       & 13.23                                      & 13.23                  & 13.23 & 13.23 & 13.23 & 13.23 & Constant  \\
            InstantNGP~\cite{instant_mller_2022} & 11.68                                      & 11.68                  & 11.68 & 11.68 & 11.68 & 11.68 & Constant  \\
            K-Planes~\cite{kplanes_2023}         & 35.66                                      & 35.66                  & 35.66 & 35.66 & 35.66 & 35.66 & Constant  \\
            MK-Planes                            & 35.66                                      & 35.96                  & 36.32 & 36.76 & 37.27 & 37.86 & Quadratic \\
            MK-Planes$^\star$                    & 35.66                                      & 35.88                  & 36.10 & 36.32 & 36.54 & 36.76 & Linear    \\
            CoGS~\cite{yu2023cogs}               & 42.24                                      & 43.23                  & 44.23 & 45.23 & 46.23 & 47.22 & Linear    \\
            \ours (Ours)                         & 34.52                                      & 34.52                  & 34.52 & 34.52 & 34.52 & 34.52 & Constant  \\
            \bottomrule
        \end{tabular}%
    }
    \label{tab:parameters}
    % \vspace{-1.5ex}
\end{table}

\boldparagraph{View Synthesis Quality Comparison on \simdata and \realdata dataset} We provide detailed quantitative results on the \simdata and \realdata datasets in~\cref{tab:supp_ominisim} and~\cref{tab:supp_interReal} provide detailed quantitative results on the \simdata and \realdata datasets, respectively. \ours outperforms prior works on most metrics and achieves the best PSNR on the \#challenging and \#office subsets with a significant margin. Note that the \#challenging and \#office subsets contain scenes with multiple interactive objects and large deformable fields, which are challenging for existing methods. We report the score as NaN if the model fails to converge or is out of memory during training multiple times.

\begin{table}[ht]
    \centering
    \setlength{\tabcolsep}{8pt}
    \renewcommand{\arraystretch}{1.0}
    \caption{\textbf{Detailed Quantitative results on \simdata Dataset}. \ours outperforms prior works on most metrics and achieves the best PSNR on the \#challenging subset with a significant margin.}
    \label{tab:supp_ominisim}
    \resizebox{1.0\columnwidth}{!}{%
        \begin{tabular}{llrrrrrrrrr}
            \toprule
            Dataset               & Metric & NeRF~\cite{mildenhall2020nerf} & Instant-NGP~\cite{instant_mller_2022} & HyperNeRF~\cite{park2021hypernerf} & K-Planes~\cite{kplanes_2023} & CoNeRF~\cite{conerf_kania_2022} & MK-Planes~\cite{kplanes_2023} & MK-Planes$^\star$~\cite{kplanes_2023} & CoGS~\cite{yu2023cogs} & \ours  \\
            \midrule
            \#seq001\_Rs          & PSNR   & 25.941                         & 25.768                                & NaN                                & 33.136                       & 34.035                          & 32.169                        & 32.092                                & 32.211                 & 34.784 \\
            \#seq001\_Rs          & SSIM   & 0.931                          & 0.933                                 & NaN                                & 0.953                        & 0.957                           & 0.946                         & 0.946                                 & 0.968                  & 0.974  \\
            \#seq001\_Rs          & LPIPS  & 0.118                          & 0.113                                 & NaN                                & 0.093                        & 0.135                           & 0.110                         & 0.110                                 & 0.068                  & 0.048  \\
            \#seq002\_Rs          & PSNR   & 28.616                         & 28.660                                & NaN                                & 34.765                       & 34.286                          & 36.532                        & 34.580                                & 34.497                 & 35.190 \\
            \#seq002\_Rs          & SSIM   & 0.950                          & 0.946                                 & NaN                                & 0.967                        & 0.951                           & 0.976                         & 0.968                                 & 0.979                  & 0.969  \\
            \#seq002\_Rs          & LPIPS  & 0.096                          & 0.112                                 & NaN                                & 0.074                        & 0.217                           & 0.036                         & 0.074                                 & 0.051                  & 0.070  \\
            \#seq003\_Ihlen       & PSNR   & 26.720                         & 28.255                                & 33.551                             & 35.217                       & 34.700                          & 34.758                        & 34.753                                & 36.816                 & 35.323 \\
            \#seq003\_Ihlen       & SSIM   & 0.940                          & 0.944                                 & 0.946                              & 0.964                        & 0.953                           & 0.966                         & 0.966                                 & 0.980                  & 0.966  \\
            \#seq003\_Ihlen       & LPIPS  & 0.120                          & 0.121                                 & 0.268                              & 0.097                        & 0.244                           & 0.087                         & 0.090                                 & 0.077                  & 0.094  \\
            \#seq004\_Ihlen       & PSNR   & 30.847                         & 31.800                                & 31.115                             & 36.157                       & 32.684                          & 34.863                        & 35.000                                & 31.055                 & 36.712 \\
            \#seq004\_Ihlen       & SSIM   & 0.927                          & 0.942                                 & 0.878                              & 0.955                        & 0.888                           & 0.919                         & 0.926                                 & 0.915                  & 0.962  \\
            \#seq004\_Ihlen       & LPIPS  & 0.104                          & 0.102                                 & 0.389                              & 0.085                        & 0.366                           & 0.145                         & 0.135                                 & 0.209                  & 0.072  \\
            \#seq005\_Beechwood   & PSNR   & 27.183                         & 27.295                                & 30.699                             & 31.944                       & 32.549                          & 33.195                        & 33.098                                & 33.664                 & 33.623 \\
            \#seq005\_Beechwood   & SSIM   & 0.930                          & 0.937                                 & 0.906                              & 0.944                        & 0.927                           & 0.961                         & 0.959                                 & 0.978                  & 0.962  \\
            \#seq005\_Beechwood   & LPIPS  & 0.127                          & 0.112                                 & 0.291                              & 0.105                        & 0.245                           & 0.076                         & 0.080                                 & 0.058                  & 0.072  \\
            \#seq006\_Beechwood   & PSNR   & 27.988                         & 28.150                                & 29.513                             & 31.861                       & 30.058                          & 31.541                        & 31.521                                & 31.272                 & 32.206 \\
            \#seq006\_Beechwood   & SSIM   & 0.938                          & 0.938                                 & 0.907                              & 0.951                        & 0.917                           & 0.951                         & 0.951                                 & 0.974                  & 0.959  \\
            \#seq006\_Beechwood   & LPIPS  & 0.103                          & 0.119                                 & 0.314                              & 0.097                        & 0.283                           & 0.095                         & 0.096                                 & 0.059                  & 0.077  \\
            \#seq007\_Beechwood   & PSNR   & 23.201                         & 22.902                                & 31.259                             & 30.979                       & 33.451                          & 30.136                        & 30.089                                & 27.367                 & 30.360 \\
            \#seq007\_Beechwood   & SSIM   & 0.885                          & 0.886                                 & 0.913                              & 0.938                        & 0.935                           & 0.942                         & 0.942                                 & 0.893                  & 0.946  \\
            \#seq007\_Beechwood   & LPIPS  & 0.220                          & 0.219                                 & 0.289                              & 0.140                        & 0.229                           & 0.120                         & 0.121                                 & 0.219                  & 0.107  \\
            \#seq008\_Benevolence & PSNR   & 25.750                         & 25.574                                & 32.691                             & 31.914                       & 34.319                          & 30.926                        & 30.916                                & 33.795                 & 33.393 \\
            \#seq008\_Benevolence & SSIM   & 0.943                          & 0.940                                 & 0.945                              & 0.948                        & 0.960                           & 0.941                         & 0.941                                 & 0.980                  & 0.970  \\
            \#seq008\_Benevolence & LPIPS  & 0.113                          & 0.123                                 & 0.229                              & 0.107                        & 0.185                           & 0.118                         & 0.116                                 & 0.072                  & 0.067  \\
            \#seq009\_Benevolence & PSNR   & 24.326                         & 24.386                                & 29.596                             & 32.836                       & 31.225                          & 31.500                        & 31.471                                & 33.205                 & 32.030 \\
            \#seq009\_Benevolence & SSIM   & 0.921                          & 0.922                                 & 0.897                              & 0.956                        & 0.932                           & 0.954                         & 0.953                                 & 0.975                  & 0.962  \\
            \#seq009\_Benevolence & LPIPS  & 0.124                          & 0.128                                 & 0.327                              & 0.090                        & 0.248                           & 0.088                         & 0.090                                 & 0.074                  & 0.071  \\
            \#seq010\_Merom       & PSNR   & 22.927                         & 22.765                                & 28.985                             & 30.120                       & 31.092                          & 29.461                        & 29.396                                & 30.254                 & 30.029 \\
            \#seq010\_Merom       & SSIM   & 0.917                          & 0.925                                 & 0.939                              & 0.960                        & 0.957                           & 0.960                         & 0.959                                 & 0.974                  & 0.966  \\
            \#seq010\_Merom       & LPIPS  & 0.173                          & 0.158                                 & 0.275                              & 0.093                        & 0.233                           & 0.087                         & 0.088                                 & 0.065                  & 0.074  \\
            \#seq011\_Merom       & PSNR   & 26.732                         & 27.077                                & NaN                                & 33.394                       & 30.483                          & 32.951                        & 32.910                                & 31.767                 & 33.426 \\
            \#seq011\_Merom       & SSIM   & 0.932                          & 0.933                                 & NaN                                & 0.959                        & 0.932                           & 0.959                         & 0.959                                 & 0.968                  & 0.960  \\
            \#seq011\_Merom       & LPIPS  & 0.112                          & 0.117                                 & NaN                                & 0.074                        & 0.246                           & 0.073                         & 0.072                                 & 0.091                  & 0.068  \\
            \#seq012\_Pomaria     & PSNR   & 26.856                         & 27.074                                & NaN                                & 35.185                       & 33.065                          & 32.248                        & 32.209                                & 37.284                 & 33.367 \\
            \#seq012\_Pomaria     & SSIM   & 0.936                          & 0.943                                 & NaN                                & 0.972                        & 0.954                           & 0.966                         & 0.966                                 & 0.985                  & 0.969  \\
            \#seq012\_Pomaria     & LPIPS  & 0.138                          & 0.126                                 & NaN                                & 0.059                        & 0.199                           & 0.075                         & 0.075                                 & 0.047                  & 0.061  \\
            \#seq013\_Pomaria     & PSNR   & 25.277                         & 24.018                                & NaN                                & 30.860                       & 33.682                          & 30.390                        & 30.299                                & 32.868                 & 33.592 \\
            \#seq013\_Pomaria     & SSIM   & 0.925                          & 0.930                                 & NaN                                & 0.943                        & 0.964                           & 0.931                         & 0.930                                 & 0.981                  & 0.970  \\
            \#seq013\_Pomaria     & LPIPS  & 0.154                          & 0.161                                 & NaN                                & 0.123                        & 0.166                           & 0.162                         & 0.164                                 & 0.045                  & 0.056  \\
            \#seq014\_Wainscott   & PSNR   & 26.011                         & 25.966                                & NaN                                & 32.517                       & 29.580                          & 30.511                        & 30.504                                & 31.885                 & 31.197 \\
            \#seq014\_Wainscott   & SSIM   & 0.927                          & 0.924                                 & NaN                                & 0.955                        & 0.925                           & 0.951                         & 0.951                                 & 0.969                  & 0.952  \\
            \#seq014\_Wainscott   & LPIPS  & 0.105                          & 0.116                                 & NaN                                & 0.077                        & 0.244                           & 0.082                         & 0.083                                 & 0.067                  & 0.083  \\
            \#seq015\_Wainscott   & PSNR   & 27.257                         & 27.191                                & NaN                                & 30.721                       & 32.307                          & 28.288                        & 28.134                                & 32.949                 & 34.266 \\
            \#seq015\_Wainscott   & SSIM   & 0.953                          & 0.951                                 & NaN                                & 0.955                        & 0.962                           & 0.942                         & 0.942                                 & 0.975                  & 0.976  \\
            \#seq015\_Wainscott   & LPIPS  & 0.080                          & 0.092                                 & NaN                                & 0.083                        & 0.202                           & 0.110                         & 0.108                                 & 0.078                  & 0.050  \\
            \#seq016\_Wainscott   & PSNR   & 21.953                         & 21.660                                & 28.364                             & 30.414                       & 30.205                          & 28.915                        & 28.710                                & 31.965                 & 29.746 \\
            \#seq016\_Wainscott   & SSIM   & 0.897                          & 0.895                                 & 0.909                              & 0.951                        & 0.935                           & 0.952                         & 0.951                                 & 0.976                  & 0.955  \\
            \#seq016\_Wainscott   & LPIPS  & 0.175                          & 0.194                                 & 0.327                              & 0.089                        & 0.260                           & 0.086                         & 0.087                                 & 0.066                  & 0.083  \\
            \#seq017\_Benevolence & PSNR   & 26.364                         & 26.367                                & 27.533                             & 29.833                       & 30.349                          & 29.254                        & 26.565                                & 28.701                 & 31.645 \\
            \#seq017\_Benevolence & SSIM   & 0.927                          & 0.920                                 & 0.897                              & 0.937                        & 0.923                           & 0.933                         & 0.887                                 & 0.970                  & 0.948  \\
            \#seq017\_Benevolence & LPIPS  & 0.128                          & 0.143                                 & 0.318                              & 0.118                        & 0.238                           & 0.119                         & 0.218                                 & 0.073                  & 0.093  \\
            \#seq018\_Benevolence & PSNR   & 28.236                         & 24.296                                & 32.551                             & 34.690                       & 34.297                          & 33.049                        & 33.002                                & 34.963                 & 34.187 \\
            \#seq018\_Benevolence & SSIM   & 0.918                          & 0.809                                 & 0.911                              & 0.951                        & 0.936                           & 0.953                         & 0.952                                 & 0.976                  & 0.958  \\
            \#seq018\_Benevolence & LPIPS  & 0.145                          & 0.342                                 & 0.293                              & 0.093                        & 0.248                           & 0.090                         & 0.091                                 & 0.114                  & 0.081  \\
            \#seq019\_Rs          & PSNR   & 20.059                         & 20.854                                & 33.119                             & 34.462                       & 34.598                          & 33.679                        & 33.653                                & 25.947                 & 35.223 \\
            \#seq019\_Rs          & SSIM   & 0.794                          & 0.808                                 & 0.950                              & 0.956                        & 0.963                           & 0.963                         & 0.962                                 & 0.879                  & 0.969  \\
            \#seq019\_Rs          & LPIPS  & 0.425                          & 0.424                                 & 0.270                              & 0.106                        & 0.225                           & 0.087                         & 0.089                                 & 0.327                  & 0.068  \\
            \#seq020\_Merom       & PSNR   & 23.273                         & 24.074                                & 31.280                             & 30.462                       & 32.580                          & 30.655                        & 30.626                                & 31.280                 & 32.869 \\
            \#seq020\_Merom       & SSIM   & 0.823                          & 0.852                                 & 0.970                              & 0.929                        & 0.914                           & 0.919                         & 0.918                                 & 0.970                  & 0.954  \\
            \#seq020\_Merom       & LPIPS  & 0.306                          & 0.259                                 & 0.086                              & 0.140                        & 0.276                           & 0.139                         & 0.142                                 & 0.086                  & 0.078  \\
            \bottomrule
        \end{tabular}
    }
\end{table}

\begin{table}[ht]
    \centering
    \setlength{\tabcolsep}{8pt}
    \renewcommand{\arraystretch}{1.0}
    \caption{\textbf{Detailed Quantitative results on \realdata Dataset}. Our method outperforms others in most settings, with a significant advantage of PSNR, SSIM, and LPIPS on the \#challenging subset.}
    \label{tab:supp_interReal}
    \resizebox{1.0\columnwidth}{!}{%
        \begin{tabular}{llrrrrrrr}
            \toprule
            dataset              & Metric & NeRF~\cite{mildenhall2020nerf} & Instant-NGP~\cite{instant_mller_2022} & HyperNeR~\cite{park2021hypernerf}F & K-Planes~\cite{kplanes_2023} & CoNeRF~\cite{conerf_kania_2022} & CoGS~\cite{conerf_kania_2022} & \ours  \\
            \midrule
            \#seq01\_transformer & PSNR   & 20.094                        & 20.619                                & 24.651                             & 26.881                       & 27.260                          & 31.067                        & 30.396 \\
            \#seq01\_transformer & SSIM   & 0.725                         & 0.805                                 & 0.638                              & 0.791                        & 0.739                           & 0.943                         & 0.912  \\
            \#seq01\_transformer & LPIPS  & 0.182                         & 0.167                                 & 0.495                              & 0.185                        & 0.355                           & 0.060                         & 0.060  \\
            \#seq02\_transformer & PSNR   & 20.093                        & 20.028                                & 24.433                             & 26.232                       & 26.917                          & 30.513                        & 29.706 \\
            \#seq02\_transformer & SSIM   & 0.736                         & 0.778                                 & 0.635                              & 0.763                        & 0.732                           & 0.938                         & 0.899  \\
            \#seq02\_transformer & LPIPS  & 0.210                         & 0.196                                 & 0.477                              & 0.223                        & 0.357                           & 0.062                         & 0.069  \\
            \#seq03\_door        & PSNR   & 20.001                        & 20.652                                & 27.144                             & 29.278                       & 29.850                          & 31.998                        & 32.709 \\
            \#seq03\_door        & SSIM   & 0.785                         & 0.831                                 & 0.878                              & 0.920                        & 0.922                           & 0.962                         & 0.960  \\
            \#seq03\_door        & LPIPS  & 0.250                         & 0.250                                 & 0.316                              & 0.101                        & 0.231                           & 0.071                         & 0.044  \\
            \#seq04\_dog         & PSNR   & 20.044                        & 20.206                                & 25.691                             & 30.350                       & 28.567                          & 32.455                        & 32.519 \\
            \#seq04\_dog         & SSIM   & 0.723                         & 0.819                                 & 0.730                              & 0.894                        & 0.815                           & 0.950                         & 0.943  \\
            \#seq04\_dog         & LPIPS  & 0.196                         & 0.178                                 & 0.435                              & 0.107                        & 0.324                           & 0.074                         & 0.049  \\
            \#seq05\_sit         & PSNR   & 21.558                        & 24.211                                & 24.944                             & 27.970                       & 26.252                          & 27.169                        & 30.161 \\
            \#seq05\_sit         & SSIM   & 0.480                         & 0.727                                 & 0.573                              & 0.773                        & 0.633                           & 0.767                         & 0.886  \\
            \#seq05\_sit         & LPIPS  & 0.178                         & 0.236                                 & 0.543                              & 0.207                        & 0.463                           & 0.232                         & 0.084  \\
            \#seq06\_stand       & PSNR   & 23.109                        & 24.483                                & 24.833                             & 27.285                       & 26.159                          & 31.442                        & 29.400 \\
            \#seq06\_stand       & SSIM   & 0.643                         & 0.699                                 & 0.574                              & 0.736                        & 0.627                           & 0.919                         & 0.868  \\
            \#seq06\_stand       & LPIPS  & 0.123                         & 0.260                                 & 0.538                              & 0.237                        & 0.470                           & 0.104                         & 0.089  \\
            \#seq07\_flower      & PSNR   & 21.150                        & 21.813                                & 25.334                             & 26.545                       & 26.854                          & 28.435                        & 28.208 \\
            \#seq07\_flower      & SSIM   & 0.721                         & 0.747                                 & 0.712                              & 0.759                        & 0.748                           & 0.893                         & 0.844  \\
            \#seq07\_flower      & LPIPS  & 0.302                         & 0.319                                 & 0.489                              & 0.321                        & 0.425                           & 0.165                         & 0.188  \\
            \#seq08\_office      & PSNR   & 21.187                        & 21.474                                & 25.188                             & 26.309                       & 26.040                          & 27.510                        & 28.663 \\
            \#seq08\_office      & SSIM   & 0.735                         & 0.743                                 & 0.714                              & 0.754                        & 0.720                           & 0.897                         & 0.848  \\
            \#seq08\_office      & LPIPS  & 0.371                         & 0.358                                 & 0.545                              & 0.341                        & 0.520                           & 0.138                         & 0.181  \\
            \bottomrule
        \end{tabular}
    }
\end{table}

\begin{figure}[t]
    \centering
    \footnotesize
    \begin{minipage}[!b]{0.28\textwidth}
        \centering
        \setlength{\tabcolsep}{2.0pt}
        \renewcommand{\arraystretch}{0.9}
        \begin{tabular}{lccccc}
            \toprule
            \multicolumn{5}{c}{Depth L1 Error $\downarrow$}                           \\
            \midrule
            scene            & CoNeRF                      & MK-Planes$^\star$ & CoGS  & \ours     \\
            \midrule
            \#\texttt{seq04} & \nd 0.029                   & 0.037             & 0.381 & \fs 0.018 \\
            \#\texttt{seq08} & \fs 0.018                   & 0.301             & 0.655 & \nd 0.025 \\
            \#\texttt{seq14} & \nd 0.042                   & 0.103             & 0.810 & \fs 0.039 \\
            \#\texttt{seq15} & \fs 0.019                   & 0.568             & 0.690 & \nd 0.021 \\
            \#\texttt{seq17} & \blackx                     & \nd 0.282         & 0.706 & \fs 0.019 \\
            \#\texttt{seq19} & \blackx                     & \nd 0.136         & 0.689 & \fs 0.034 \\
            \#\texttt{avg}   & \nd \textcolor{gray}{0.027} & 0.238             & 0.655 & \fs 0.026 \\
            \bottomrule
        \end{tabular}
    \end{minipage}
    \hfill
    \begin{minipage}[!b]{0.51\textwidth}
        \centering
        \includegraphics[width=\textwidth]{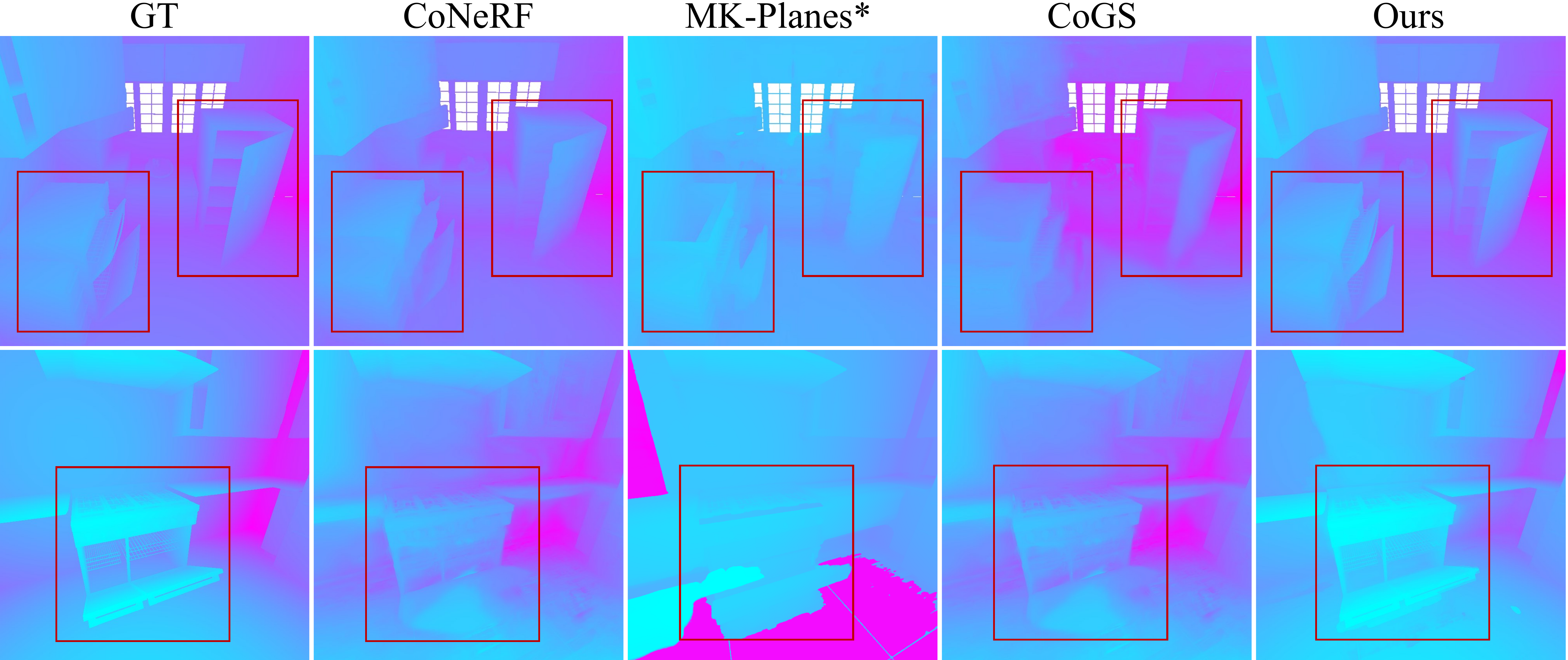}
    \end{minipage}
    \vspace{-1ex}
    \caption{\textbf{Structure Reconstruction Performance on \simdata Dataset.} Our method surpasses most previous works on chosen subsets.}
    \label{fig:depth_l1_supp}
    \vspace{-2ex}
\end{figure}

\boldparagraph{Interactive Scenes Geometry Comparison}. To evaluate the completeness of the topological structure of interactive objects, we employ the depth L1 error metric. As shown in \cref{fig:depth_l1_supp}, our method outperforms SOTA methods on scenes from \simdata. While existing methods excel in RGB image rendering, they struggle with depth structure representation. Specifically, CoNeRF~\cite{conerf_kania_2022} performs relatively well in \#seq08 and \#seq15 but fails in large-scale scenes (\#seq17 and \#seq19).  CoGS~\cite{yu2023cogs} exhibit notable artifacts in the depth map. Moreover, MK-Planes$^\star$ also fails to recover depth around interactive objects. In contrast, our method achieves the lowest depth error and renders satisfying depth maps, demonstrating its accurate interactive scene modeling capabilities.

\begin{figure}[t]
    % \vspace{-2ex}
    \centering
    \footnotesize
    \begin{minipage}[!b]{0.45\textwidth}
        \centering
        \setlength{\tabcolsep}{2.0pt}
        \renewcommand{\arraystretch}{0.9}
        \begin{tabular}{cclcccc}
            \hline
            \multicolumn{6}{c}{mIOU $\uparrow$}                                                                                                                         \\
            \hline
            \multicolumn{3}{c}{setting}                                    & SAM~\cite{kirillov2023segany} & LERF~\cite{lerf_kerr_2023} & Ours                          \\
            \multirow{4}{*}{\rotatebox[origin=l]{90}{\scriptsize\simdata}} &                               & \#easy                     & \nd 61.58 & 23.60 & \fs 86.94 \\
                                                                           &                               & \#medium                   & \nd 55.13 & 19.40 & \fs 86.32 \\
                                                                           &                               & \#challenging              & \nd 63.86 & 19.87 & \fs 90.41 \\
                                                                           &                               & \#avg                      & \nd 59.11 & 21.74 & \fs 86.86 \\
            \midrule
            \multirow{3}{*}{\rotatebox[origin=l]{90}{\scriptsize\realdata}} &         & \#medium                   & 93.27 & 27.63 & 84.37          \\
                                                                            &         & \#challenging              & 91.50 & 34.39 & \fs 91.90 \\
                                                                            &         & \#avg                      & 92.82 & 29.32 & 86.26          \\
            \hline
        \end{tabular}
    \end{minipage}
    \hfill
    \begin{minipage}[!b]{0.5\textwidth}
        \centering
        \includegraphics[width=\textwidth]{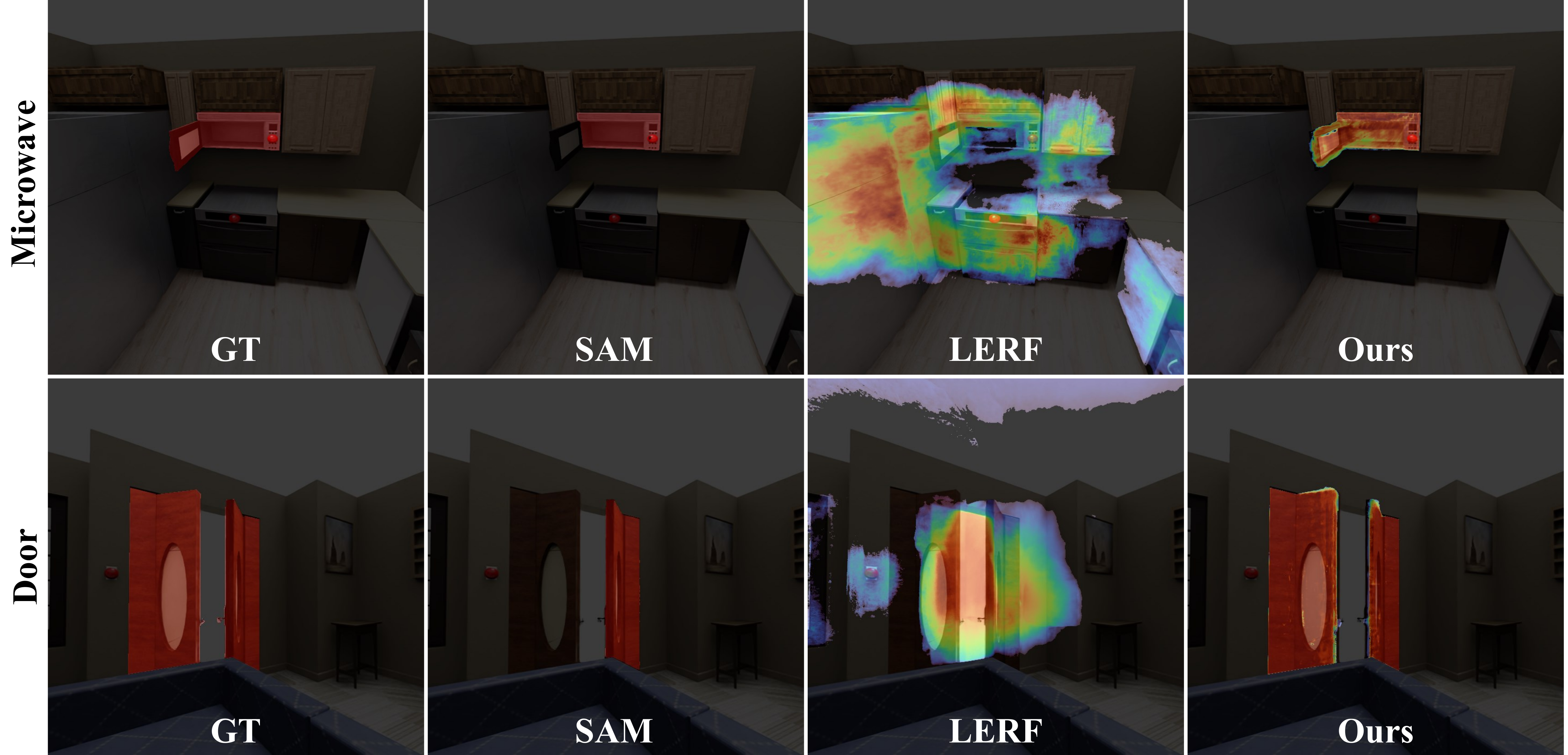}
    \end{minipage}
    \vspace{-1ex}
    \caption{\textbf{Language Grounding Performance on \simdata Dataset.} left): Our method gains the highest mIOU score. right): LiveScene's grounding exhibits clearer boundaries than other methods.}
    \label{fig:miou_supp}
    % \vspace{-3ex}
\end{figure}

\boldparagraph{Language Grounding Comparison}. We assess the language grounding performance on \simdata dataset using mIOU metric. \cref{fig:miou_supp} suggests that our method obtains the highest mIOU score, with an average of 86.86. In contrast, traditional methods like LERF~\cite{lerf_kerr_2023} encounter difficulties in locating objects precisely, with an average mIOU of 21.74. Meanwhile, 2D methods like SAM~\cite{kirillov2023segany} fail to accurately segment the whole target under specific viewing angles, as objects appear discontinuous in the image. Conversely, our method perceives the completeness of the object and has clear knowledge of its boundaries, demonstrating its advantage in language grounding tasks.

\begin{figure}[t]
    \vspace{-1ex}
    \begin{center}
        \includegraphics[width=1\linewidth]{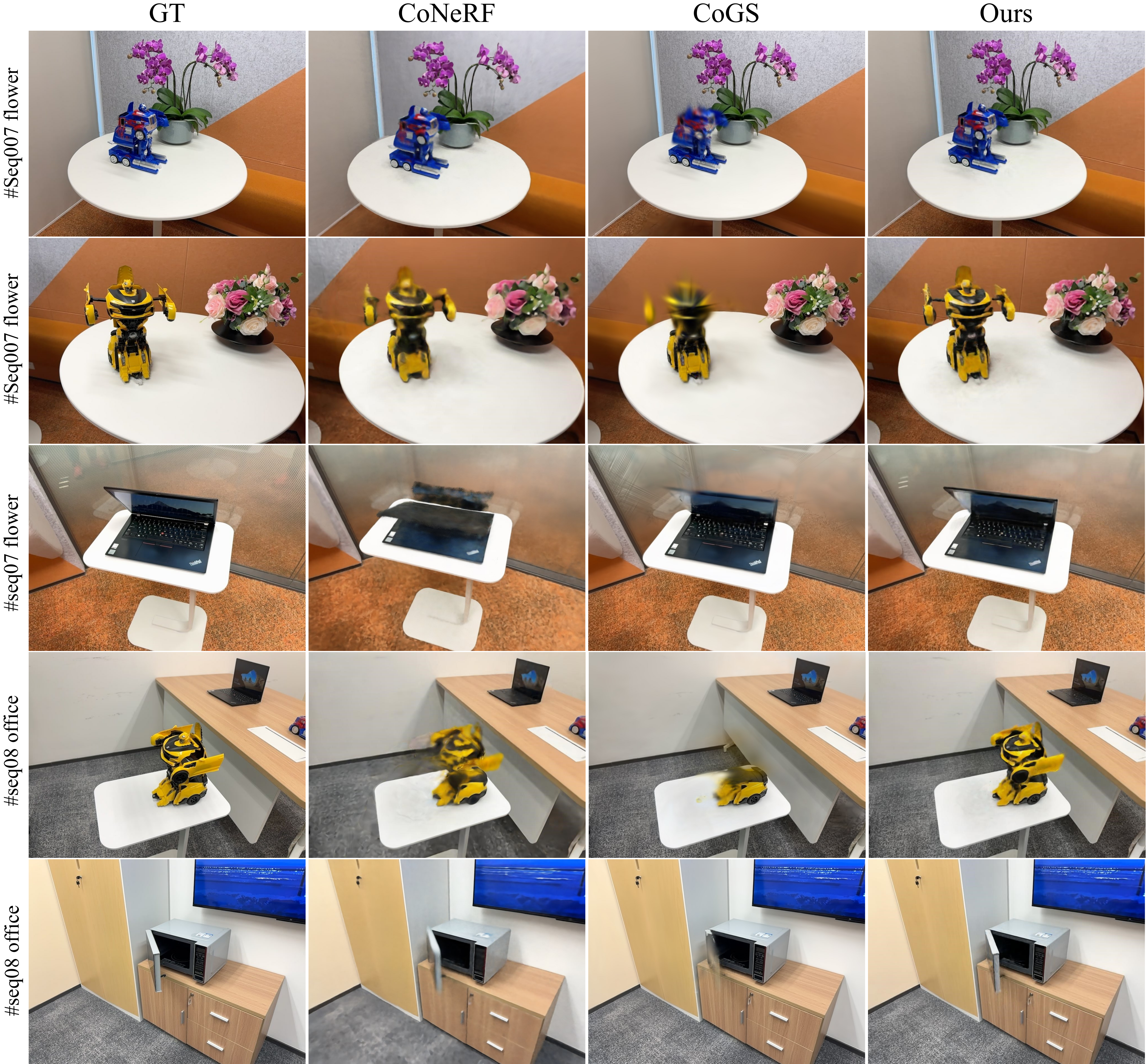}
    \end{center}
    \vspace{-1ex}
    \caption{\textbf{View Synthesis Visualization on \realdata Dataset}.  We compare our method with SOTA methods on RGB rendering across real scenes. \ours obtained more detailed and accurate representations of the objects. While other methods fail to capture the object's shape and cause significant artifacts.}
    \label{fig:supp_render_real}
    \vspace{-2ex}
\end{figure}

\begin{figure}[t]
    \vspace{-1ex}
    \begin{center}
        \includegraphics[width=1\linewidth]{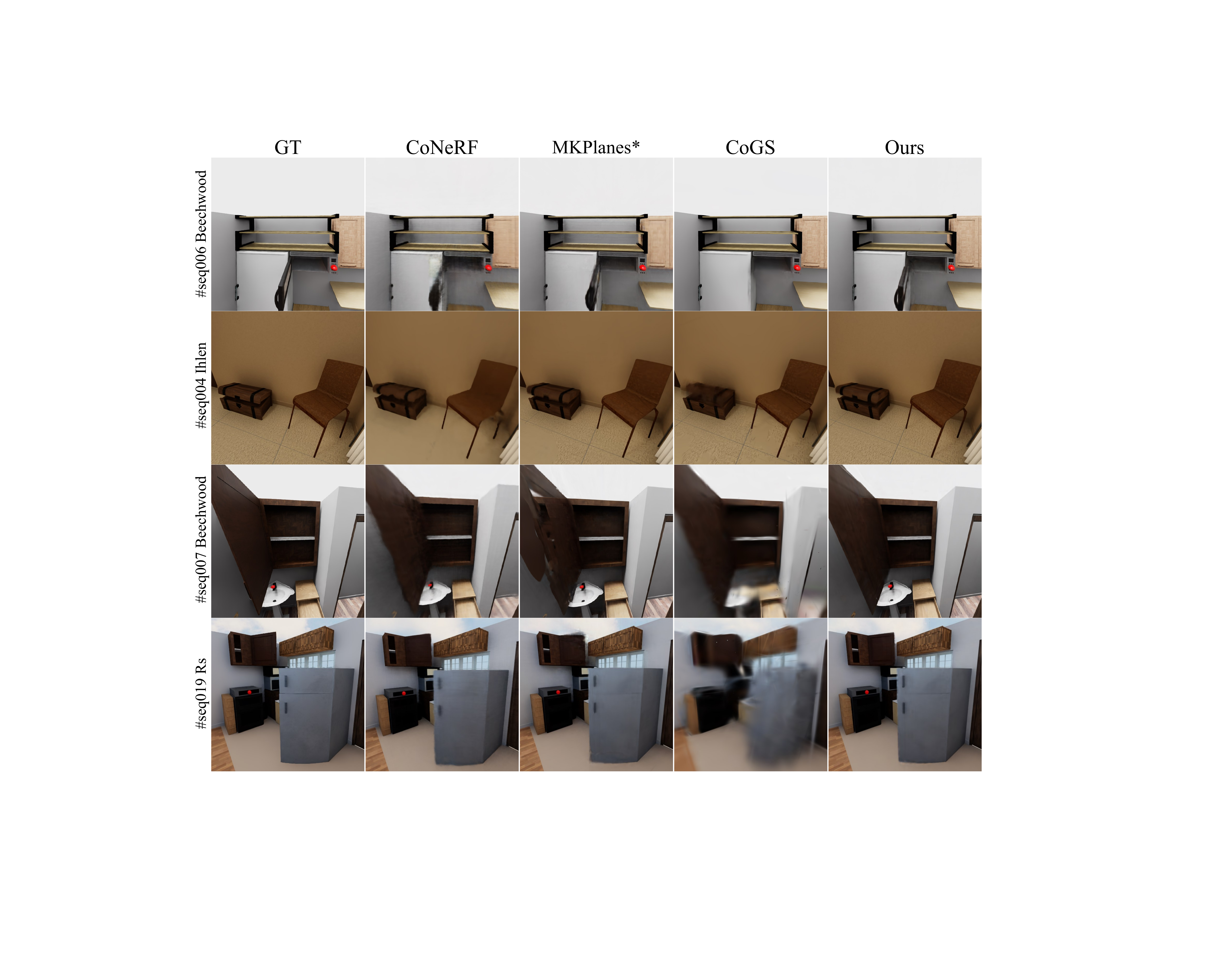}
    \end{center}
    \vspace{-1ex}
    \caption{\textbf{View Synthesis Visualization on \realdata Dataset}. compared with the other methods, \ours reconstructs clear and accurate object shapes and textures.}
    \label{fig:supp_render_sim}
    \vspace{-2ex}
\end{figure}

\begin{figure}[t]
    \vspace{-1ex}
    \begin{center}
        \includegraphics[width=1\linewidth]{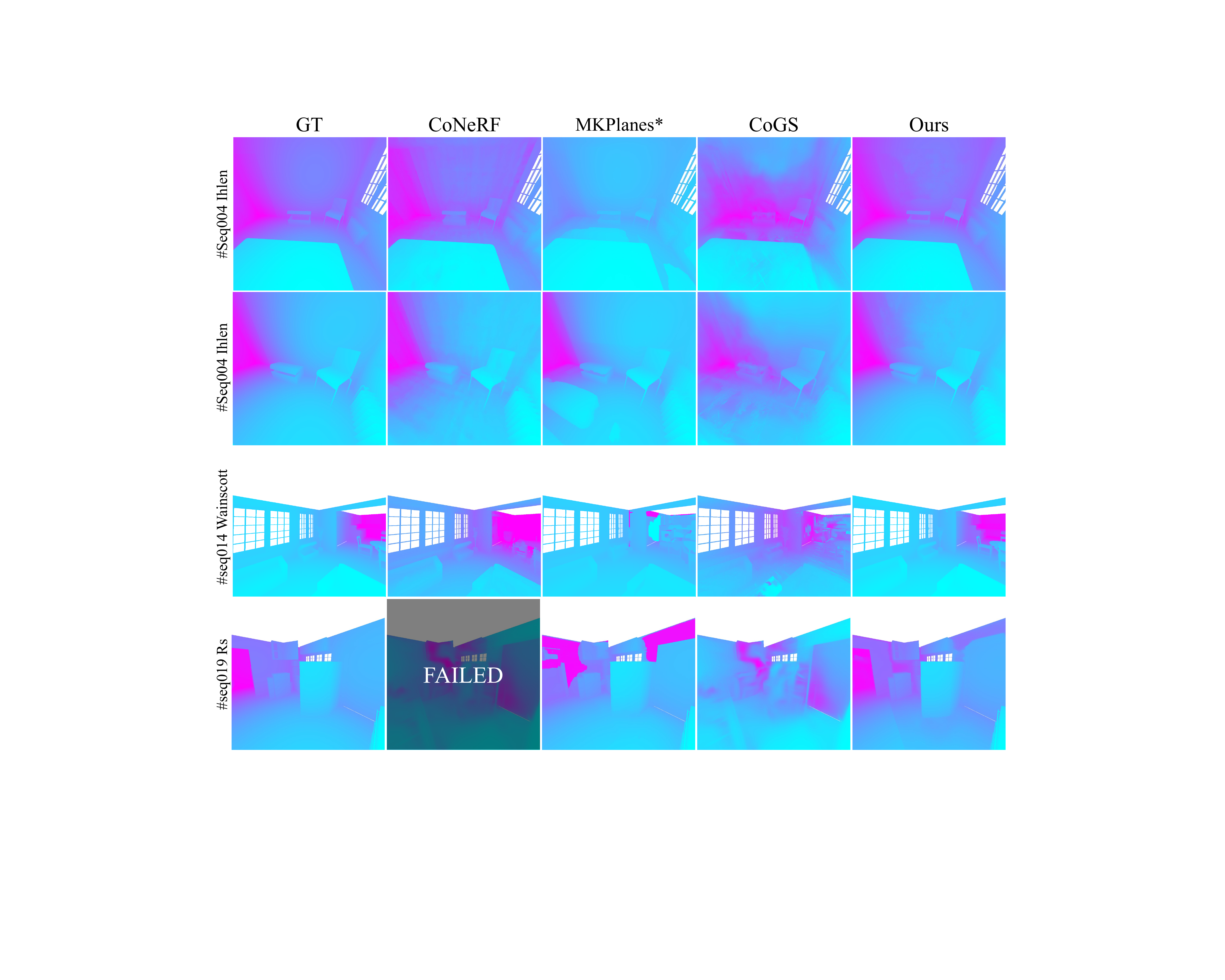}
    \end{center}
    \vspace{-1ex}
    \caption{Illustration of the depth map comparison on the \simdata datasets. Our method can generate more accurate depth maps than other methods, demonstrating the effectiveness of interactive scene reconstruction. In contrast, other methods either fail to capture the object's shape or cause significant artifacts.}
    \label{fig:supp_depth}
    \vspace{-2ex}
\end{figure}

\begin{figure}[t]
    \vspace{-1ex}
    \begin{center}
        \includegraphics[width=1\linewidth]{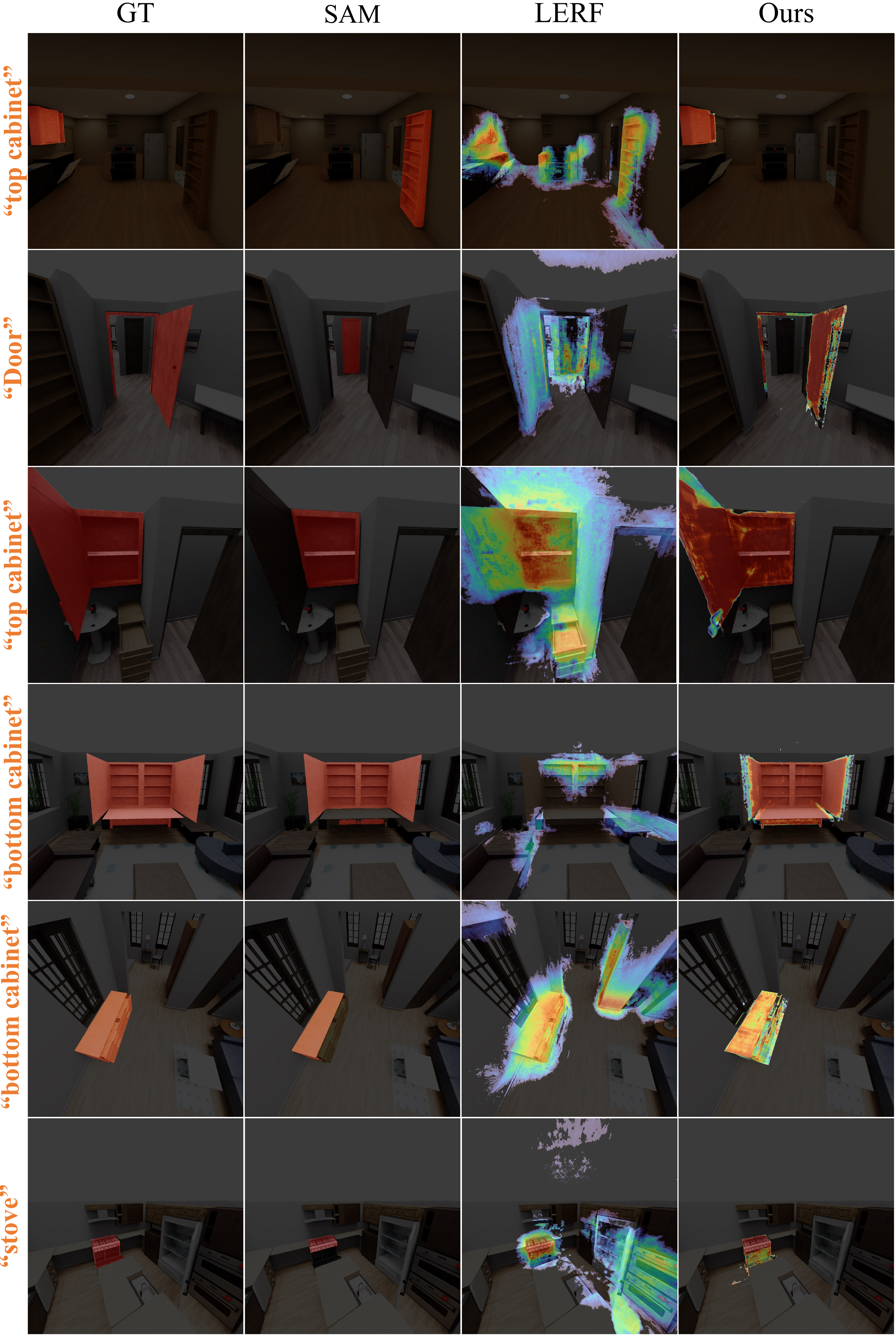}
    \end{center}
    \vspace{-1ex}
    \caption{Illustration of the language grounding comparison on the \simdata datasets. Compared to LeRF, our method can locate more accurate interactive objects, overcoming the obvious inconsistency problem in interactions, while maintaining accurate boundaries. In contrast, LeRF suffers from a diffusion phenomenon in object localization due to changes in object topology structure.}
    \label{fig:supp_lang}
    \vspace{-2ex}
\end{figure}

\boldparagraph{More Detailed Rendering Comparison}
We provide more detailed visual comparisons, including RGB, depth, and language grounding on the \simdata and \realdata datasets in \cref{fig:supp_render_real}, \cref{fig:supp_render_sim}, \cref{fig:supp_depth}, and \cref{fig:supp_lang}, respectively. Our method surpasses existing approaches by reconstructing more detailed and accurate representations of the objects. In both datasets, \ours can generate more accurate and detailed object shapes and textures, especially for scenes with multiple interactive objects and large deformable fields. Compared with LERF~\cite{lerf_kerr_2023}, our method can generate more accurate language grounding results, which is crucial for interactive object manipulation tasks, demonstrated in \cref{fig:supp_lang}.

\subsection{More Ablation Studies}
\begin{figure}[t]
    \vspace{-4ex}
    \begin{center}
        \includegraphics[width=1.0\linewidth]{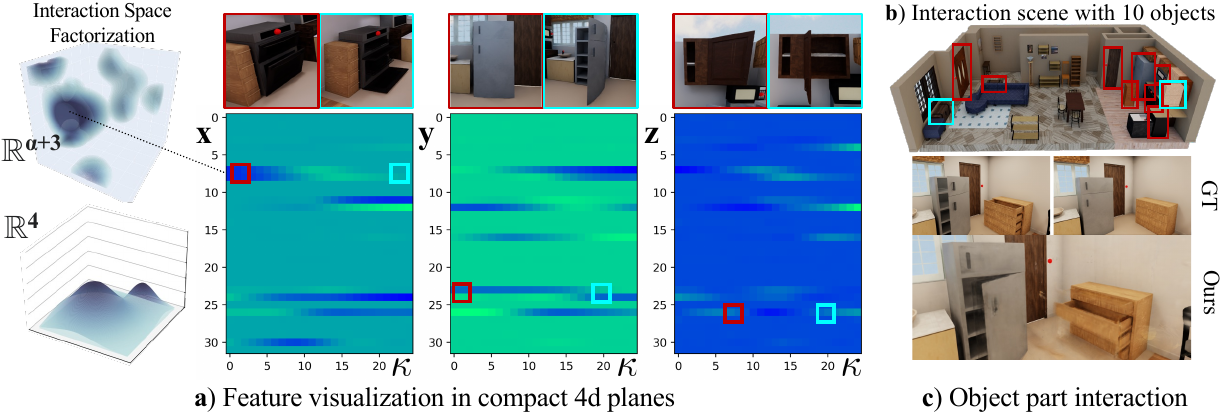}
    \end{center}
    \vspace{-1ex}
    \caption{\textbf{a)} Visualization of x-$\kappa$, y-$\kappa$ and z-$\kappa$ latent feature planes. \textbf{b)} Sce ne with more objects. \textbf{c)} Part-level interaction.}
    \label{fig:feature}
    \vspace{-2ex}
\end{figure}

\boldparagraph{4D Deformable Feature Visualization.} We provide additional interaction feature visualization of x-z, y-z, and z-k in~\cref{fig:feature}(a) to illustrate latent feature distribution. It can be seen that the features are clustered around the spatial coordinates of interactive objects, corresponding to the local deformable fields in Sec 3.2 of the manuscript. \cref{fig:feature}(b) validates the performance of \ours in scenarios with up to 10 complex interactive objects. Notably, our method demonstrates robustness in rendering quality, which does not degrade significantly as the object number increases. The number of objects is not a major limiting and our method is still feasible as long as the dataset provides mask and control variable labels. In contrast, the occlusion and topological complexity between objects do affect the reconstruction results, which will be discussed in the limitations section. In~\cref{fig:feature}(c), we demonstrate the fine-grained control capability of \ours on a refrigerator and cabinet dataset without part-based labels. Our method can control a part of the object even though there are no individual part-based interaction variable labels. However, the effect is not entirely satisfactory, due to the lack of labels and CLIP's limited understanding of spatial relationships.

\begin{figure}[t]
    \vspace{-1ex}
    \begin{center}
        \includegraphics[width=1\linewidth]{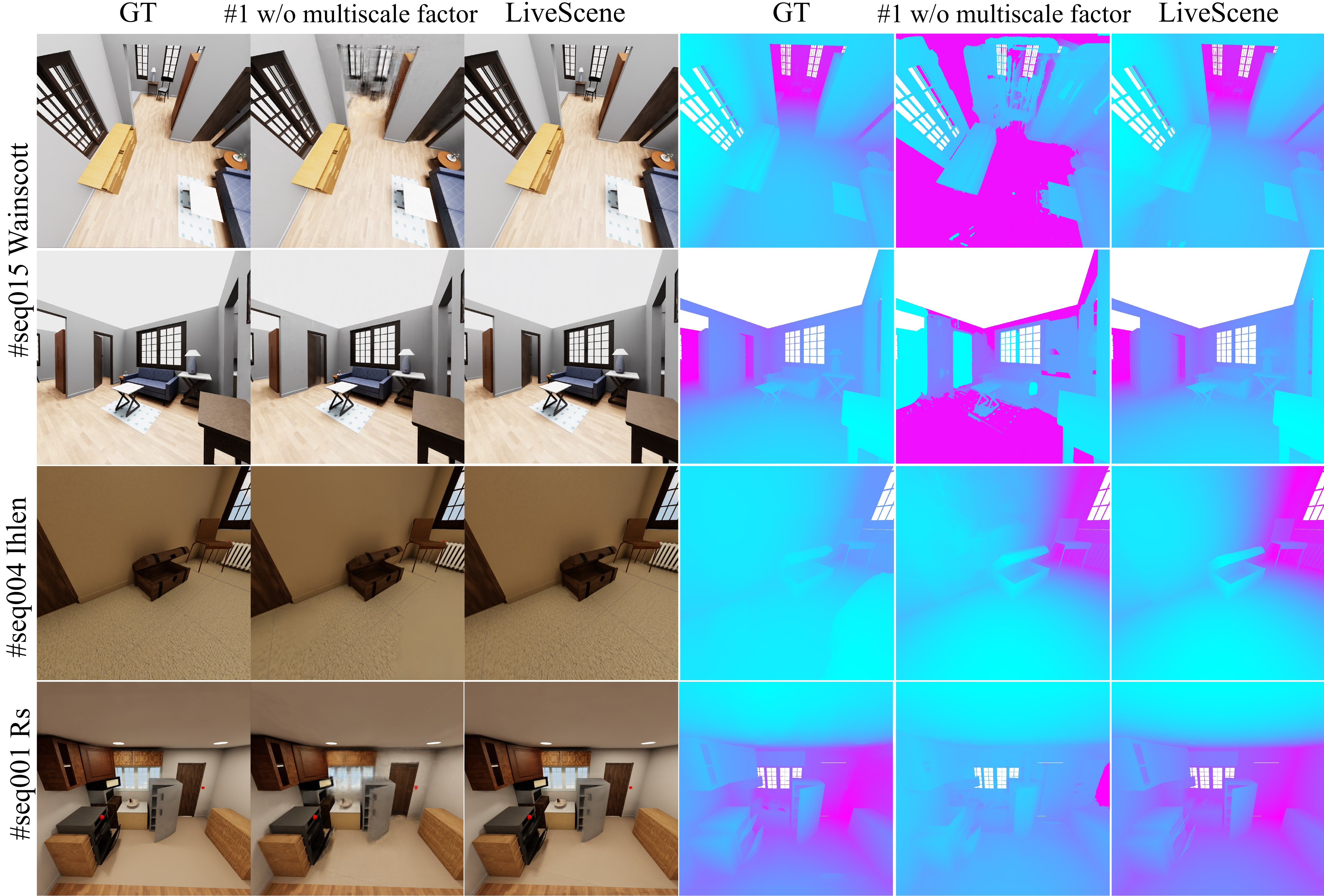}
    \end{center}
    \vspace{-1ex}
    \caption{More ablation of multi-scale factorization on the \simdata dataset. We compare the performance of \ours with w/o multiscale factor. The results show that the multi-scale factorization can improve the model's performance by capturing the object's detailed structure and texture. However, the model without multi-scale factorization performs poorly in depth rendering, illustrating the improvements of multi-scale factorization in scene geometric modeling.}
    \label{fig:supp_multiscale}
    \vspace{-2ex}
\end{figure}

\boldparagraph{Ablation Study on Multi-scale Factorization.} We conduct more ablation studies on the \simdata dataset to evaluate the effectiveness of the multi-scale factorization. The results show that the multi-scale factorization can improve the model's performance by capturing the object's detailed structure and texture. However, the model without multi-scale factorization performs poorly in depth rendering, illustrating the improvements of multi-scale factorization in scene geometric modeling. The results are shown in \cref{fig:supp_multiscale}.

\begin{figure}[t]
    \vspace{-1ex}
    \begin{center}
        \includegraphics[width=1\linewidth]{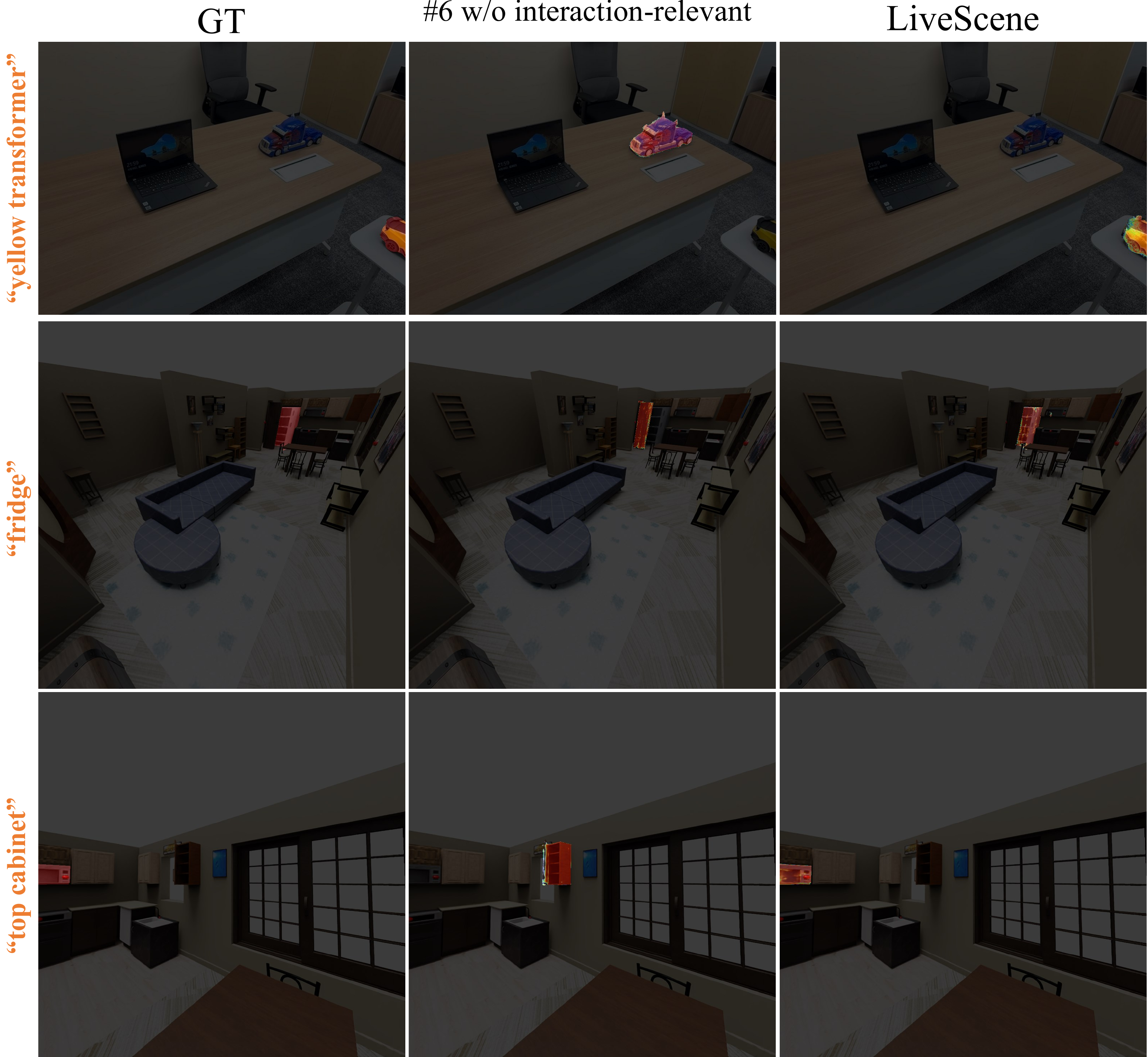}
    \end{center}
    \vspace{-1ex}
    \caption{More ablation of interactive object modeling on the \simdata dataset. We compare the performance of \ours with w/o interaction-aware language embedding. The results show that the interaction-aware language embedding can effectively improve the model's performance when encouraging significant scene topological changes. While the model without interaction-aware language embedding fails to ground the correct object because of the lack of interaction-aware information.}
    \label{fig:supp_consistency}
    \vspace{-2ex}
\end{figure}

\boldparagraph{Ablation Study on Interaction-aware Language Embedding.} We conduct more ablation studies on the \simdata dataset to evaluate the effectiveness of the interaction-aware language embedding. The results show that the interaction-aware language embedding can effectively improve the model's performance when encouraging significant scene topological changes. While the model without interaction-aware language embedding fails to ground the correct object because of the lack of interaction-aware information. The results are shown in \cref{fig:supp_consistency}.

\begin{figure}[t]
    \vspace{-1ex}
    \begin{center}
        \includegraphics[width=1\linewidth]{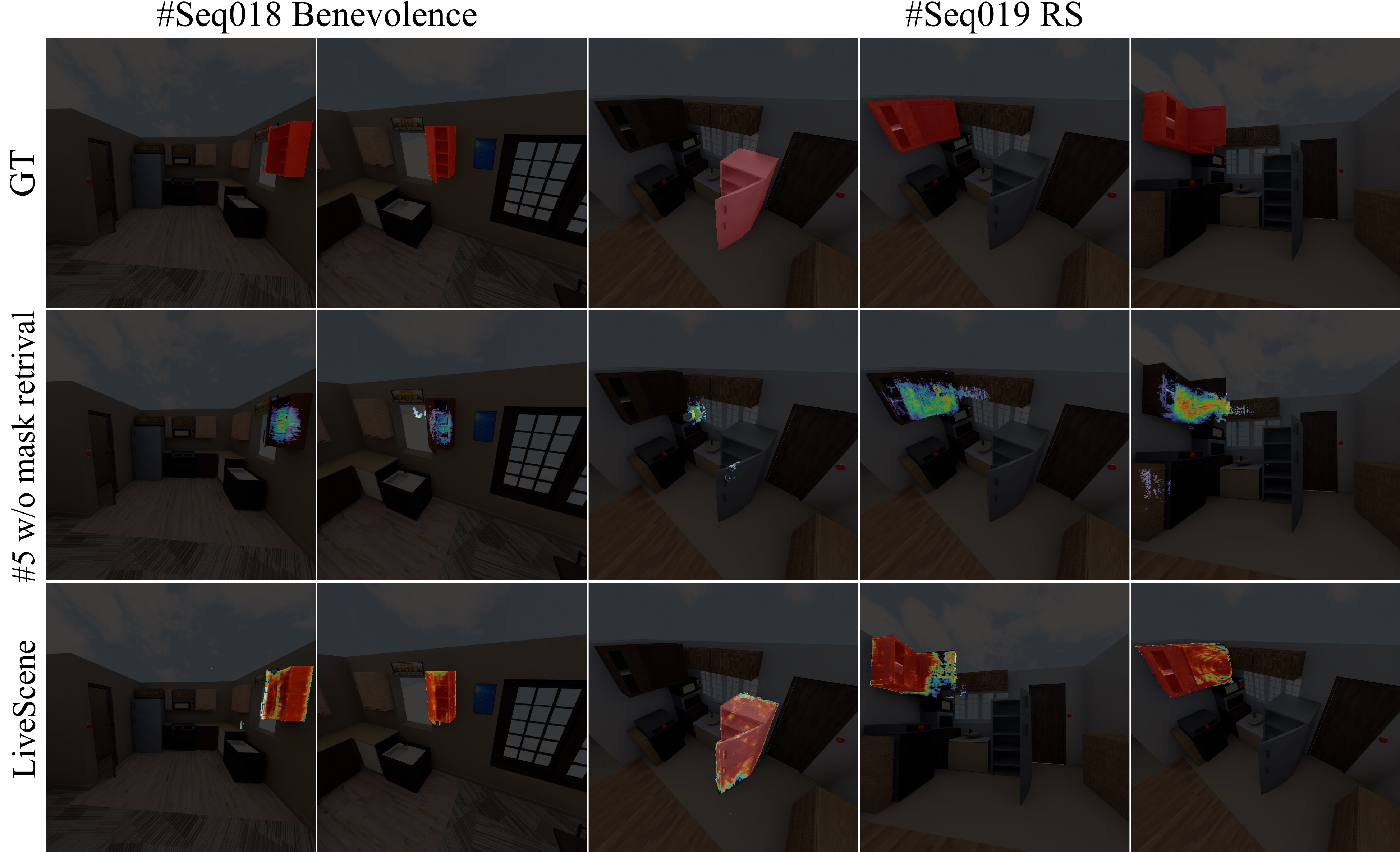}
    \end{center}
    \vspace{-1ex}
    \caption{By applying the proposed multiscale factor and maximum probability embedding retrieval, the model achieves better performance with higher storage efficiency and training speed, and the grounding results will also be more concentrated in the object region. The fundamental reason is that this method decouples language from the 3D scene to the object level, rather than the entire 3D space.}
    \label{fig:supp_maskretrival}
    \vspace{-2ex}
\end{figure}

\boldparagraph{Maximum Probability Embeds Retrieval.} We conduct more ablation studies on the \simdata dataset to evaluate the effectiveness of the maximum probability embedding retrieval. The results show that the maximum probability embedding retrieval can improve the model's performance with higher storage efficiency and training speed, and the grounding results will also be more concentrated in the object region. The fundamental reason is that this method decouples language from the 3D scene to the object level, rather than the entire 3D space. The results are shown in \cref{fig:supp_maskretrival}.
 
\clearpage
\medskip
{
    \small
    \bibliographystyle{plain}
    \bibliography{main}
}
\end{document}